\documentclass[runningheads]{llncs}

\usepackage{eccv}

\usepackage{eccvabbrv}

\usepackage{graphicx}
\usepackage{amsmath,amssymb} 
\graphicspath{{figures/}}
\usepackage{booktabs}
\usepackage{booktabs}
\usepackage{listings}
\usepackage{courier} 

\usepackage[accsupp]{axessibility}  

\usepackage{hyperref}

\usepackage{orcidlink}

\begin{document}

\title{Moiré Video Authentication: A Physical Signature Against AI Video Generation}

\titlerunning{Moiré Video Authentication}


\author{Yuan Qing\thanks{Equal contribution} \and
Kunyu Zheng\protect\footnotemark[1] \and
Lingxiao Li\protect\footnotemark[1] \and
Boqing Gong \and
Chang Xiao}

\authorrunning{Y.~Qing et al.}

\institute{Boston University, Boston, USA\\
\email{\{ymqing, kunyuz, lxli, bgong, xchang\}@bu.edu}\\
\small{\url{https://yuanqing-ai.github.io/physical_video_signature/}}}

\maketitle

\begin{abstract}
  Recent advances in video generation have made AI-synthesized content increasingly difficult to distinguish from real footage. We propose a physics-based authentication signature that real cameras produce naturally, but that generative models cannot faithfully reproduce. Our approach exploits the Moir\'e effect: the interference fringes formed when a camera views a compact two-layer grating structure. 
  We utilize the \emph{Moir\'e motion invariant} as a verification criterion: under real image formation, fringe phase and grating image displacement are linearly coupled by optical geometry, independent of viewing distance and grating structure.
  A verifier extracts both signals from video and tests their correlation. We validate the invariant on both real-captured and AI-generated videos from multiple state-of-the-art generators, and find that real and AI-generated videos produce significantly different correlation signatures, suggesting a robust means of differentiating them. Our work demonstrates that deterministic optical phenomena can serve as physically grounded, verifiable signatures against AI-generated video.
\end{abstract}


\section{Introduction}
\label{sec:intro}

In February 2024, a finance worker at a multinational firm was tricked into transferring \$25 million after attending a video call in which every other participant, including the company's chief financial officer, were impersonated by AI-generated video~\cite{chen2024deepfake}. In October 2025, an AI-generated video mimicking a news broadcast falsely depicted an Irish presidential candidate announcing her withdrawal from the race, spreading across social media for hours before being removed~\cite{liggett2025deepfake}. Spreading unauthenticated misinformation by generative AI is no longer a hypothetical scenario; it is the new reality. Generative video models such as Sora~\cite{brooks2024sora}, Runway~\cite{runway2025gen45}, and Seedance~\cite{bytedance2026seedance} now produce footage so convincing that human viewers, automated detectors, and even forensic analysts struggle to distinguish it from real recordings~\cite{croitoru2024deepfake}. As AI-generated video increasingly spreads misinformation and erodes public trust, the need for a reliable, theoretically grounded mechanism to prove that a video is \emph{authentic} has become more urgent than ever.

Existing AI video detection broadly fall into two categories. \emph{Post-hoc} forensic detectors analyze pixel-level artifacts, temporal inconsistencies, or learned statistical fingerprints to flag AI-generated content~\cite{rossler2019faceforensics,wang2020cnn,gragnaniello2021gan}. While effective against earlier generators, these methods are locked in an arms race: each new model eliminates the artifacts that detectors exploit, demanding constant retraining and offering limited lasting guarantees. \emph{Digital watermarking} approaches, such as C2PA metadata~\cite{c2pa2022spec} and SynthID~\cite{deepmind2023synthid}, embed provenance information into media at creation time. However, metadata can be stripped, re-encoded, or forged, and digital watermarks degrade under common post-processing operations. Crucially, neither paradigm provides an unforgeable physical link between the recorded scene and the resulting video.

We propose a fundamentally different approach: a \textbf{physical authentication signature}. A real video is produced by an optical system governed by physical laws, whereas generative models learn statistical correlations from data and rarely model the underlying physics. A physical phenomenon whose appearance is tightly coupled to precise optical geometry therefore serves as a signature that real cameras produce naturally but that current video generative models cannot precisely reproduce.

\begin{figure}[t]
    \centering
    \includegraphics[width=0.99\linewidth]{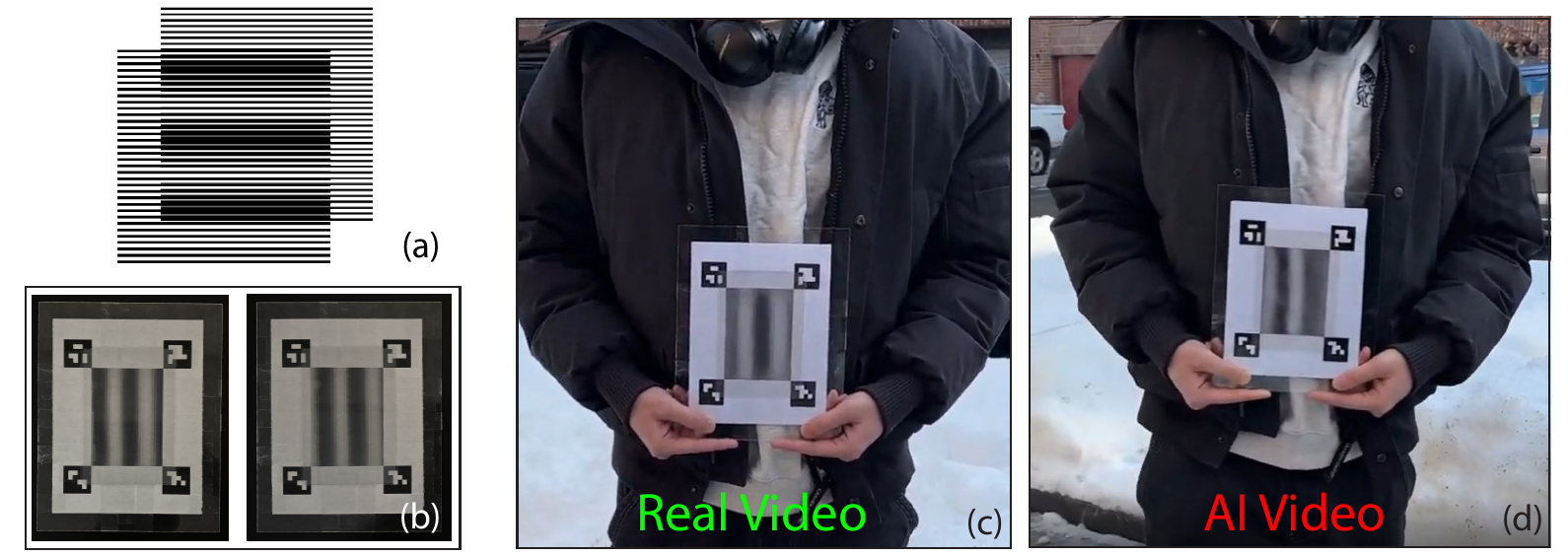}
    \caption{(a) The Moiré effect, created by overlaying two repetitive layers. (b) Our prototype Moiré signature assembly. Viewing the assembly from slightly different camera angles (left vs. right) causes the fringes to shift in phase according to deterministic physical laws. (c) In a real video, Moiré fringes appear naturally and shift predictably with camera movement. (d) In an AI-generated video, the fringes distort, and their phase shifts do not adhere to the underlying physics.}
    \label{fig:moire}
\end{figure}

We instantiate this idea through the Moir\'e effect~\cite{amidror2009moire}, an intriguing and widely observed yet often overlooked optical phenomenon. Moir\'e effect are the interference fringes that appear when two fine periodic structures, such as line gratings, are overlaid (\cref{fig:moire}-a). These fringes are highly sensitive to the precise geometric relationship between the camera and the grating assembly: as the camera moves, the fringes shift according to deterministic optical laws (\cref{fig:moire}-b). This tight coupling between camera motion and fringe motion forms a physics-based signature that real optical systems produce naturally (\cref{fig:moire}-c) but that current video generation models cannot reliably synthesize (\cref{fig:moire}-d).

The core assumption underlying our approach is that current video generation models are not grounded in physical simulation. They learn to produce plausible-looking frames from large-scale data, but they do not model the wave-optics interactions that govern Moir\'e fringe formation and evolution. Accurately reproducing the effect would require solving the exact optical problem for every frame: tiny shifts in camera position produce measurable fringe displacement, and the mapping from camera motion to fringe motion follows a strict mathematical relationship derived from first principles~\cite{amidror2009moire}. This stands in sharp contrast to the statistical interpolation that current generative models perform. Even if future models were trained on abundant footage containing Moir\'e patterns, learning the visual appearance of fringes can hardly, if not impossibly, equip a model to solve the underlying optical equations.

Leveraging this insight, we propose a practical authentication system that works as follows. A compact grating assembly, consisting of a lenticular sheet mounted above a printed stripe pattern, is placed in the scene (\eg, worn by a speaker at a press conference, see \cref{fig:moire-vs-verilight}). Given a recorded video with relative movement between the assembly and the camera (which occurs naturally from the speaker's body motion even when the camera is static) our algorithm (1) extracts the Moir\'e fringe pattern and maps it to a canonical coordinate frame, (2) tracks the fringe phase changes over time, and (3) computes the correlation between the observed fringe phase and the displacement of the assembly in the video. A high correlation certifies authenticity; a low or absent correlation flags the video as potentially synthetic. Importantly, analogous to digital watermarking, our method provides positive authentication for videos that contain the signature; it does not make claims about videos lacking one.

This paper contributes a novel idea supported by theoretical grounding and empirical evidence for validating the feasibility. Our contributions are:
\begin{itemize}
    \item \textbf{A novel Moir\'e-based video authentication framework.} We introduce the concept of using the Moir\'e effect as a passive, physics-based authentication signature. 
    We utilize the Moir\'e motion invariant as a verification criterion:
    fringe phase displacement and grating image displacement are linearly coupled; the resulting correlation is independent of viewing position and grating structure, yielding a verifiable signal extractable from video alone.
    \item \textbf{A practical implementation and feasibility validation.} We build a proof-of-concept system using off-the-shelf materials (a lenticular sheet, printed patterns, and ArUco markers) and demonstrate that the physics-predicted correlation holds in real captured video, confirming the practical viability of the approach.
    \item \textbf{A threat model analysis.} We systematically examine how our method performs against a spectrum of attack scenarios, characterizing the conditions under which the Moir\'e signature remains robust.  We also synthesize a large number of AI-generated videos in our best effort to reproduce the Moir\'e pattern, and the results show that even cherry-picked and carefully engineered generated videos fail to achieve correlation comparable to that of real video.
\end{itemize}

\section{The Moir\'e Effect}
\label{sec:moire}

A Moir\'e pattern is a large-scale interference pattern that emerges when two fine periodic structures are superimposed with a slight difference in their spatial frequencies~\cite{amidror2009moire}. Moir\'e patterns appear ubiquitously in everyday life, from the shimmering bands seen through overlapping fences to the rippling fringes visible when a digital display is photographed at certain positions, and they have found broad application in optical metrology~\cite{xiao2021moireboard,gabrielyan2007moire,campos2024moirewidgets}, mechanical deformation analysis~\cite{amidror2009moire,takasaki1970moire,sciammarella1982moire,kafri1990moire}, and computational art~\cite{hersch2004moire,sethapakdi2025fabobscura}.

In our system, the Moir\'e pattern is generated by a compact structure consisting of two layers of line gratings with slightly different spatial periods, separated by a thin gap of thickness $g$. Because the gap causes each viewing position to reveal a different relative alignment of the two layers, the superposition produces prominent Moir\'e fringes whose positions are extremely sensitive to camera-grating geometry. The fringes act as a \emph{geometric amplifier}: a microscopic change in viewing position produces a macroscopic fringe shift, yielding a visually prominent response that is easily captured on video yet difficult for generative models to synthesize correctly.

Without loss of generality, we restrict our analysis to \textbf{one-dimensional Moir\'e patterns} (parallel line fringes), which is the configuration used in our implementation. However, two-dimensional Moir\'e patterns formed by crossed or rotated gratings~\cite{amidror2009moire, xiao2021moireboard} can be easily implemented and extended in the same principle. As a feasibility study, in this paper we only consider the 1D scenario.

\subsection{Theory}
\label{sec:moire:theory}
Our optical derivation builds on classical Moir\'e theory, as established in prior textbooks~\cite{amidror2009moire,takasaki1970moire,kafri1990moire}.
We illustrate the relationship between camera motion and Moir\'e fringe displacement for the one-dimensional two-layer grating structure. 
Consider two aligned gratings with spatial periods $p_f$ (front) and $p_r$ (rear). From standard Moir\'e theory, their superposition produces fringes with beat period $p_m = p_f p_r / |p_f - p_r|$. A lateral relative shift $\delta$ between the layers displaces the fringes by $|\Delta x_m| = M |\delta|$, where $M = p_f / |p_f - p_r|$ is the \textbf{Moir\'e magnification factor}.

In our structure the layers are rigidly fixed; the apparent shift $\delta$ arises from viewing-angle parallax. We analyze motion along the axis $x$ perpendicular to the grating lines (the only direction that produces fringe shifts). A camera at distance $D$ undergoing a lateral displacement $\Delta x_c$ sees an apparent layer shift $\delta = g \, \Delta x_c / D$, where $g$ denotes the gap between the two grating layers. Converting fringe displacement to \textbf{unwrapped phase} ($\Delta\phi = 2\pi \Delta x_m / p_m = 2\pi \delta / p_r$, where the $p_f$ and $|p_f - p_r|$ terms cancel) and substituting $\delta$ yields
\begin{equation}
\label{eq:phase_full}
\Delta \phi = \pm \frac{2\pi g}{p_r \cdot D} \cdot \left(1 + \frac{g}{D}\right) \cdot \Delta x_c,
\end{equation}
where the sign depends on coordinate convention and the factor $(1 + g/D)$ accounts for the rear grating's projected period on the front grating plane.

\subsection{Our Insight: The Correlation Invariant}
\label{sec:moire:invariant}

The preceding relationship shows that Moir\'e fringe displacement depends on camera motion. We now show that this same quantity governs the apparent motion of the grating structure itself in the camera image, yielding a correlation invariant that is independent of the unknown distance $D$. We first derive the invariant for pure camera translation, then generalize to arbitrary camera motion that includes rotation.

In our physical setting, the grating gap $g$ is on the order of millimeters while the observation distance $D \geq 0.5$\,m, so the factor $(1 + g/D)$ in \cref{eq:phase_full} is negligible. We therefore approximate \cref{eq:phase_full} as
\begin{equation}
\label{eq:phase_approx}
\Delta \phi = \pm \frac{2\pi g}{p_r \cdot D} \cdot \Delta x_c.
\end{equation}

\paragraph{Pure translation.}
Under a pinhole camera model with focal length $f$, a pure translational camera displacement $\Delta x_c$ along the perpendicular axis at distance $D$ from the grating structure causes the structure's centroid to shift in the image plane by
\begin{equation*}
\Delta u_{\text{trans}} = \pm f \cdot \frac{\Delta x_c}{D},
\end{equation*}
where the sign depends on coordinate convention. Combining this with \cref{eq:phase_approx}, we eliminate the common factor $\Delta x_c / D$ to obtain
\begin{equation}
\label{eq:phase_trans}
\Delta \phi = \pm \frac{2\pi g}{p_r \cdot f} \cdot \Delta u_{\text{trans}}.
\end{equation}

\paragraph{General camera motion.}
In practice, camera motion is a combination of translation and rotation. Under the pinhole model, a small camera rotation $\Delta\theta$ about the axis parallel to the grating lines (\ie pan or tilt that displaces the structure image along the perpendicular axis $x$) produces an additional image displacement $\Delta u_{\text{rot}}$. The total observed image displacement of the structure is therefore
\begin{equation*}
\Delta u = \Delta u_{\text{trans}} + \Delta u_{\text{rot}}.
\end{equation*}
Critically, rotation does \emph{not} induce parallax between the two grating layers. A pure rotation shifts both layers identically in the image, producing zero apparent relative displacement ($\delta = 0$) and hence zero fringe phase change. Only the translational component $\Delta u_{\text{trans}}$ generates parallax and drives $\Delta\phi$. Therefore \cref{eq:phase_trans} holds in the general case, with $\Delta u_{\text{trans}} = \Delta u - \Delta u_{\text{rot}}$.

For an authentic video, the temporal sequences of fringe phase changes $\{\Delta\phi_t\}$ and translational image displacements $\{\Delta u_{\text{trans},t}\}$ are linearly coupled with near-unity absolute correlation:
\begin{equation}
\label{eq:correlation}
|\rho| = \left|\text{Corr}\bigl(\{\Delta\phi_t\},\ \{\Delta u_{\text{trans},t}\}\bigr)\right| \approx 1.
\end{equation}
We call this the \textbf{Moir\'e motion invariant}. The proportionality constant need not be known to the verifier: $D$ has cancelled entirely, and although $f$, $g$, and $p_r$ may be unknown, they are fixed for a given camera and grating structure. Because correlation measures linear coupling regardless of the slope's value, a verifier simply extracts both sequences from the pixel data and checks whether they are linearly coupled, without requiring any calibration or knowledge of the scene geometry.

In general, isolating $\Delta u_{\text{trans}}$ from the observed image displacement $\Delta u$ requires estimating and subtracting the rotational component $\Delta u_{\text{rot}}$, which can be done purely from the video, for instance by tracking distant background features or decomposing camera pose via multi-point correspondence methods such as Perspective-n-Point (PnP) pose estimation~\cite{zhang2000flexible}. In our implementation, we perform this rotation compensation using the four ArUco fiducial markers already present on the grating assembly as 3D-to-2D correspondences for PnP-based camera pose recovery, allowing us to isolate the translational displacement component. We describe our concrete implementation in \cref{sec:implementation}.

In an AI-generated video, no physical grating structure exists; the generator must synthesize fringe patterns frame by frame from learned statistics. Because the generator does not solve the optical equation, the synthesized fringe motion will not maintain the strict linear coupling to the structure's position required by the invariant. Even small deviations, a fringe that shifts too fast, too slow, or in the wrong direction by a fraction of a cycle, will depress $|\rho|$ far below 1. This theoretical gap between physics-governed and statistically generated fringe behavior is the foundation of our authentication algorithm.

\section{Proof-of-Concept Implementation}
\label{sec:implementation}

We now describe our proof-of-concept system for validating the Moir\'e motion invariant. We first detail the physical grating assembly, then present the algorithmic pipeline that extracts both the fringe phase signal and the grating assembly displacement signal from a recorded video and computes their correlation.

\subsection{Physical Setup}
\label{sec:implementation:setup}

\begin{figure}[t]
    \centering
    \includegraphics[width=0.99\linewidth]{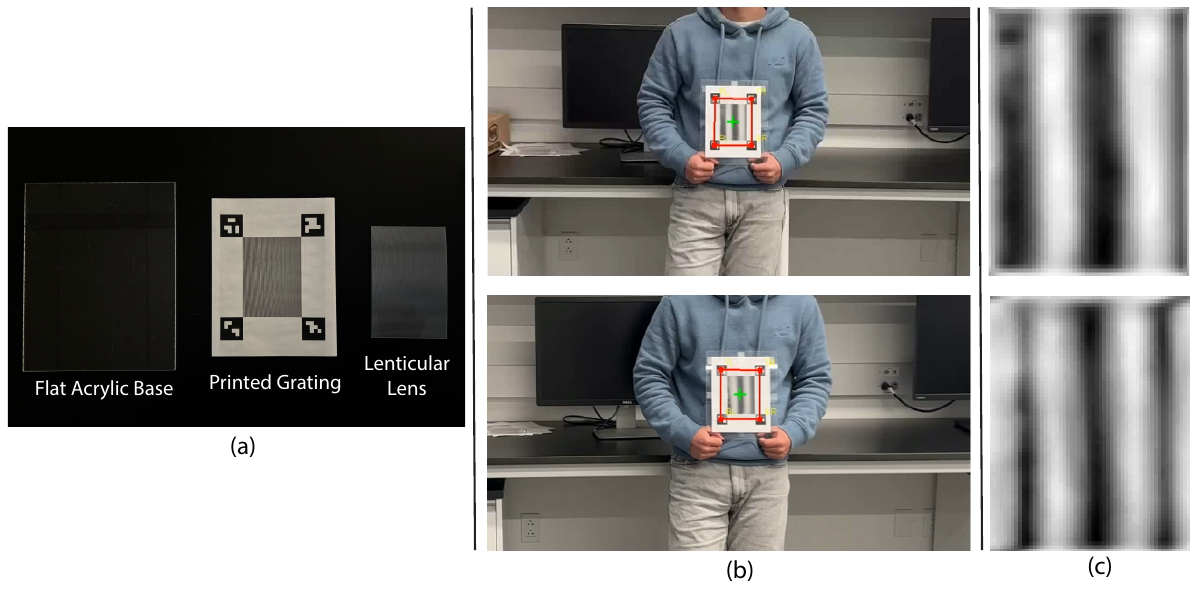}
    \caption{(a) Exploded view of the grating assembly, showing its three constituent layers: a flat acrylic base, a printed grating, and a lenticular lens. (b) Two distinct video frames illustrating our algorithm detecting the ArUco markers to isolate the Moiré fringe region. (c) The extracted Moiré fringes after a canonical transformation, corresponding to the frames in (b). The phase shift of the fringes between the two frames is clearly visible.}
    \label{fig:setup}
\end{figure}

Our grating assembly consists of three layers bonded together into a flat, rigid structure (see \cref{fig:setup}-a):

\begin{itemize}
    \item \textbf{Front layer (viewing grating).} A 1\,mm-thick lenticular lens sheet (50\,LPI, period $p_f = 0.508$\,mm) whose cylindrical lenses act as angular apertures, so that small lateral camera movements change the visible portion of the rear layer and shift the Moir\'e fringes.
    \item \textbf{Rear layer (reference grating).} A line pattern printed on paper with pitch $p_r = 0.52$\,mm. The slight mismatch between $p_f$ and $p_r$ produces Moir\'e fringes with a beat period of $p_m = p_f \cdot p_r / |p_f - p_r| \approx 22$\,mm, yielding a Moir\'e magnification factor of $M = p_f / |p_f - p_r| \approx 42\times$.
    \item \textbf{Base layer.} A flat acrylic sheet that serves as the structural substrate, holding all layers in alignment.
\end{itemize}

The three layers are assembled and attached to a flat backing surface. For ease of prototyping, we place four ArUco fiducial markers (from the \texttt{DICT\_4X4\_50} dictionary) at the corners of the grating assembly to simplify bounding-box extraction and perspective correction. Any robust localization method could replace them in a production system.

\subsection{Verification Pipeline}
\label{sec:implementation:pipeline}

Given an input video containing the grating assembly, our pipeline extracts two independent temporal signals -- the fringe phase and the grating assembly displacement -- and tests their linear coupling. The pipeline proceeds in three stages.

\paragraph{Stage 1: Grating Assembly Tracking and Fringe Extraction.}
We localize the grating assembly in each frame by detecting and tracking the four ArUco corner markers, yielding per-frame corner coordinates and a centroid trajectory $\bar{c}^{(t)}$. Using the four corners, we compute a per-frame homography that warps the grating assembly region into a canonical, front-parallel rectangle and apply CLAHE~\cite{zuiderveld1994clahe} to enhance fringe visibility. We then determine the fringe orientation via contour analysis~\cite{suzuki1985contours}, rotate the image to align fringes vertically, and collapse the 2D image into a 1D intensity profile by averaging along the fringe direction. \cref{fig:setup}-(b, c) shows the tracked Moiré region and the corresponding warped canonical frames from our pipeline.

\paragraph{Stage 2: Phase Tracking.}
We extract the fringe phase from the 1D profile using a one-dimensional FFT. Because the profile is approximately sinusoidal at the Moir\'e beat frequency, the phase of the corresponding Fourier component directly encodes the lateral position of the fringe pattern. On the first frame, we identify the dominant frequency peak and lock its bin $k^*$; on each subsequent frame we read the phase at $k^*$:
\begin{equation*}
\phi^{(t)} = \arg\!\left[\mathcal{F}\{I^{(t)}\}(k^*)\right].
\end{equation*}
To avoid $2\pi$ discontinuities, we track the phase incrementally: we wrap each frame-to-frame difference to $[-\pi, \pi]$ using $\mathrm{wrap}(\theta) = ((\theta + \pi) \bmod 2\pi) - \pi$ and accumulate the deltas into a smooth cumulative phase signal:
\begin{equation*}
\Phi^{(t)} = \sum_{\tau=1}^{t} \mathrm{wrap}\!\left(\phi^{(\tau)} - \phi^{(\tau-1)}\right).
\end{equation*}

\paragraph{Stage 3: Displacement Estimation and Correlation.}
As derived in \cref{sec:moire:invariant}, camera rotation does not induce parallax between the grating layers and therefore does not drive fringe phase change; only the translational component of camera motion does. The raw centroid trajectory $\bar{c}^{(t)}$ from Stage~1 conflates both components, so we decompose it using PnP-based pose estimation. The four ArUco marker centers provide four coplanar 3D-to-2D point correspondences per frame, which we pass to a PnP solver (\ie, OpenCV's \texttt{cv::solvePnP}) to recover the camera pose $(R_i, t_i)$ for each frame. We use a pinhole camera model with approximate intrinsics computed from image size ($f_x = f_y = \alpha \max(W, H)$). We then isolate the translation-only displacement by re-projecting the 3D centroid of the marker layout using each frame's camera position but the reference frame's orientation, yielding the rotation-compensated translational displacement $\Delta u_{\text{trans}}^{(t)}$. Because the Moir\'e fringes vary along one axis only, we project $\Delta u_{\text{trans}}^{(t)}$ onto the fringe-sensitive direction (perpendicular to the fringe lines) to obtain the scalar displacement signal.

We test the Moir\'e motion invariant (\cref{eq:phase_trans}) by computing the Pearson correlation between the cumulative phase signal $\{\Phi^{(t)}\}$ and the projected translational displacement $\{\Delta u_{\text{trans}}^{(t)}\}$. We evaluate the correlation over sliding windows of 30 frames, producing a per-window correlation curve across the video that serves as the authenticity signal. In the next section, we show that real and AI-generated videos produce markedly different correlation curves, suggesting ways to differentiate them.

\section{Validation}
\label{sec:experiments}

We validate the Moir\'e motion invariant across three categories of video: real recordings, physics-based renderings, and AI-generated videos. We apply the same verification pipeline described in \cref{sec:implementation} to all categories and compare the resulting correlation signals.

\subsection{Real Recorded Video}
\label{sec:experiments:real}

We record videos across 12 subjects in 29 distinct scenes (25 indoor, 4 outdoor) using an iPhone 15 at 1080p, 60\,fps. Each recording falls into one of three motion configurations: (1) grating assembly static with camera moving, (2) grating assembly moving with camera static, and (3) both moving simultaneously (We use notation such as "Outdoor 1" to indicate an outdoor video recorded under the first condition, and so on.). The case where both are static is excluded, as our verification requires relative movement between the camera and the grating assembly to produce fringe shifts. Across subjects, we vary the movement speed and viewing distance (approximately 1.0 to 3.0\,m). This yields a total of 87 videos, each approximately 15 seconds in duration.

\subsection{Physics-based Rendering Video}
\label{sec:experiments:pbr}

To validate the pipeline under fully controlled conditions, we reproduce the grating assembly in Blender~\cite{blender2025} and render videos with the Cycles physically-based rendering engine. The virtual assembly replicates the two-layer geometry described in \cref{sec:implementation}, and the virtual camera matches the iPhone 15's focal length and 1080p resolution. Because Cycles traces light through the two grating layers, it accurately reproduces the parallax-driven fringe formation, providing an ideal-case for verifying the pipeline.

Each rendered sequence places the grating assembly at the world origin, sized to match the physical prototype. The camera begins at $(0.5, 0, z)$ and translates linearly to $(0, 0, z)$ over 120 frames, with small random rotations injected along the trajectory to simulate realistic camera shake. We constrain the rotations so that the grating assembly and all four ArUco markers remain fully visible in every frame. We sweep the camera distance $z$ from 1.1 to 1.7\,m in 0.1\,m increments, generating 10 sequences per distance with distinct random rotation profiles, for a total of 70 rendered videos (Rendering 1 in \cref{fig:results}). We also created a group for pure translational motion without rotation (Rendering 2 in \cref{fig:results}). Additional experiments with different embodiments, pattern sizes, and motions involving 3D slant are provided in the supplementary material.

\subsection{AI-generated Video}
\label{sec:experiments:ai}

We evaluate AI-generated video as a potential attack against the Moir\'e motion invariant, testing three state-of-the-art video generation models: Veo 3.1~\cite{deepmind2025veo}, Grok Imagine~\cite{xai2025grok} (closed-source), and LTX-2~\cite{hacohen2026ltx2} (open-source). We first attempted text-to-video (T2V) generation, providing descriptions of a person holding a grating assembly with visible Moir\'e fringes (replicating our real video settings). For each of the three motion configurations, we engineered specific text prompts through 15 to 20 rounds of iterative refinement. Across all models and prompt variations, none produced output containing a recognizable or trackable Moir\'e pattern. The generated videos either omitted the fine periodic structure or rendered it as a static, incoherent texture bearing no resemblance to real Moir\'e fringes, which can be identified easily by the naked eye.

We therefore focus our quantitative evaluation on the stronger image-to-video (I2V) setting, where the attacker provides one real frame of the grating assembly as conditioning input along with a text prompt describing the desired motion. We extracted the initial frame from each of the 87 authentic videos to serve as a visual condition and refined three corresponding motion prompts over 15 to 20 rounds. For each extracted frame, we use three optimized motion prompts and generate multiple outputs, yielding 321 videos in total. Because this first frame is captured from a real camera, it contains authentic Moir\'e fringes, giving the generator a visual prior on fringe appearance.

From the initial pool of 321 synthetic videos, manual inspection revealed that the generative models produced severe temporal inconsistencies, such as the sudden disappearance or extreme deformation of the grating assembly. We therefore screened and discarded 229 such corrupted samples, yielding a final curated dataset of 92 high-quality synthetic videos that look authentic to the human eye (see the supplementary video for examples of failed AI-generated videos).

Even in the I2V setting, most current video generation models fail to reproduce the basic visual structure of the grating assembly or the ArUco fiducial markers, and such videos would be immediately flagged as non-authentic. To stress-test the robustness of the Moir\'e signature itself, we adopt a deliberately generous evaluation protocol: we manually select the corner of Moir\'e region from the first generated frame, track it using Lucas-Kanade optical flow~\cite{bouguet2001lucas} across subsequent frames, and manually correct the tracking when it deviates. This ensures the extracted signal reflects purely the generated Moir\'e pattern, independent of preprocessing failures.

\subsection{Results}
\label{sec:experiments:results}

\begin{figure}[t]
    \centering
    \includegraphics[width=\linewidth]{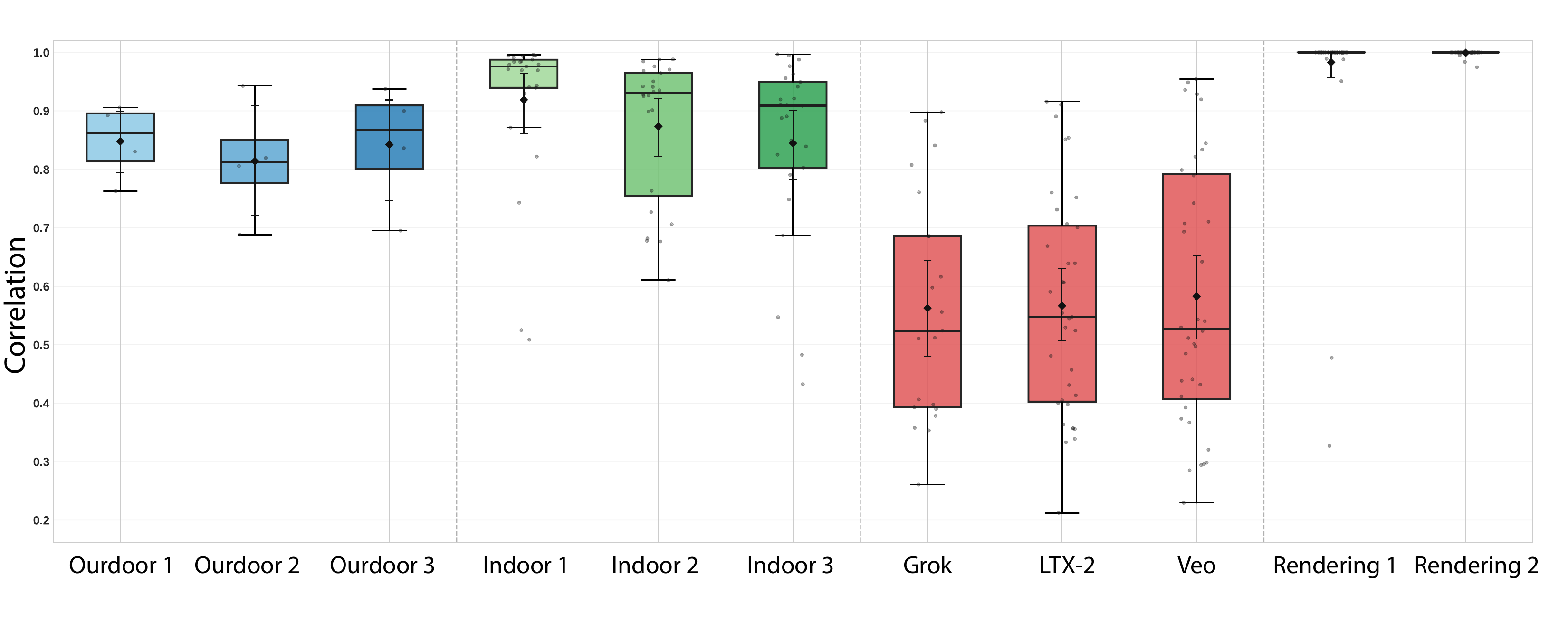}
    \caption{Distribution of Pearson correlation coefficients $|\rho|$ across all video categories. Real recordings (indoor and outdoor) cluster at high correlation, physics-based renderings achieve near-perfect correlation, and AI-generated videos concentrate at markedly lower values.}
    \label{fig:results}
\end{figure}

\cref{fig:results} shows the distribution of Pearson correlation coefficients $|\rho|$ across all video categories. Real recordings cluster at high correlation values ($\mu = 0.87$, $\sigma = 0.14$), confirming that the Moir\'e motion invariant holds under diverse real-world conditions including varying lighting, distance, and motion configurations. Physics-based renderings achieve near-perfect correlations ($\mu = 0.99$), validating that the pipeline correctly captures the underlying optical relationship under ideal conditions. In contrast, AI-generated videos concentrate at substantially lower values ($\mu = 0.57$, $\sigma = 0.21$).

To test whether the difference is statistically significant, we apply a Welch's $t$-test, which is appropriate here because the two groups have unequal variances ($\sigma = 0.14$ vs.\ $0.21$) and the sample sizes ($n = 86$ real, $n = 92$ AI-generated) are large enough for the Central Limit Theorem to ensure robustness to non-normality. The test confirms a highly significant separation ($t(160) = 11.6$, $p < 10^{-20}$). To quantify effect size, we compute Cohen's $d = 1.71$, well above the $0.8$ threshold conventionally considered a large effect. The gap is consistent across all three generators: Grok ($\mu = 0.56$), LTX-2 ($\mu = 0.57$), and Veo ($\mu = 0.58$) each fall far below the real-video baseline, with no model achieving a meaningful advantage. 

Moreover, this results should be read as a conservative lower bound shaped by deliberate choices on both sides of the experiment. On the attack side, our protocol grants every possible advantage: image-conditioned generation with a real first frame, manual Moir\'e region tracking with human correction, and cherry-picked outputs representing the best case for each model. Under a less generous protocol the attacker would also need to fool ArUco marker detection and automated Moir\'e tracking. Also, the I2V attack is itself inherently constrained, as the attacker must first obtain a real recording of the target scene to extract a conditioning frame, significantly limiting creative freedom compared to unconstrained T2V generation. On the defense side, our prototype uses a minimal, proof-of-concept pipeline. Our tracking pipeline could be improved to reduce the influence of factors such as reflections and blur in the Moiré assembly, and to use better methods for more accurate rotation estimation. \textbf{The key takeaway is therefore not the absolute threshold, but the existence of a physically grounded, measurable signal that distinguishes real from AI-generated video}. 

\subsection{Other Threat Models}
\label{sec:experiments:other_threats}

Beyond the generative attacks evaluated above, we consider two additional threat scenarios through conceptual analysis. These threats do not require new experiments but highlight important boundaries and extensions of our approach.

\paragraph{Threat: Moir\'e region splicing.}
The attacker takes a clip of an authentic grating assembly from one real video and composites it into a different (potentially AI-generated) video. The fringes in the spliced region are physically correct because they originate from a real recording.

\textbf{Analysis.} The spliced fringes carry the phase evolution from the \emph{source} recording, but our pipeline measures correlation against the grating assembly displacement in the \emph{target} video. These two signals reflect different camera-assembly geometries, so they will not be linearly related. The transplanted phase signal cannot match a different motion trajectory, and the correlation will be low by construction. We omit empirical validation for this threat because the outcome is self-evident from the formulation of the invariant.

\paragraph{Threat: Face swap and localized editing outside the Moir\'e region.}
The attacker takes a real video of person~A holding a grating assembly and applies a face-swap model (\eg DeepFaceLab~\cite{liu2023deepfacelab}, FaceFusion~\cite{facefusion2024}) to replace only the face region, making it appear as person~B. The grating assembly region is left entirely untouched.

\textbf{Analysis.} This is the most challenging threat for our system. Because the grating assembly region is unmodified, the Moir\'e fringes retain their physically correct phase evolution, and the Moir\'e motion invariant holds. Our correlation check will report high $|\rho|$, authenticating the video. The attacker obtains a video of ``person~B'' apparently holding person~A's grating assembly. Since the editing occurs entirely outside the Moir\'e region, the Moir\'e correlation check alone cannot detect this manipulation.

However, this class of attack is well addressed by a mature body of work on deepfake and face-swap detection~\cite{rossler2019faceforensics,wang2020cnn,gragnaniello2021gan,croitoru2024deepfake}. We view these as complementary: our method authenticates whether the video was produced by a physical optical system, while editing detectors identify localized tampering. The two can be combined in a layered verification pipeline. We additionally outline two forward-looking mitigations:

\textbf{Mitigation 1: Personalized Moir\'e identity (Moir\'e ID).} If the grating assembly encodes a unique, verifiable identity, analogous to a public key, the identity encoded in the assembly belongs to person~A, not person~B. This can be achieved by fabricating the grating with specific spatial frequencies that produce a unique Moir\'e beat period, acting as a frequency-domain fingerprint. Alternatively, the grating can incorporate non-uniform line spacing so the fringe pattern at different spatial positions encodes distinct identifiers. The verifier then checks not only that the Moir\'e motion invariant holds but also that the assembly identity matches the claimed identity of the person in the video.

\textbf{Mitigation 2: Spatial overlap verification protocol.} In interactive settings such as video conferencing, the platform can enforce a protocol requiring the participant to move the grating assembly across their face (\eg ``please wave your Moir\'e signature in front of your face now''). This ensures the grating assembly region spatially overlaps with the face region for at least a brief interval. Any face-swap model operating on this footage must either (a) modify the overlapping grating assembly pixels, which breaks the Moir\'e motion invariant, or (b) leave the assembly intact and distort the face around it, producing visible artifacts.

\subsection{Discussion}
\label{sec:experiments:discussion}

Our threat analysis reveals a clear division that aligns with the current landscape of video forensics. Localized editing attacks such as face swaps fall outside the scope of what Moir\'e correlation alone can detect, since the grating assembly region remains untouched. However, existing deepfake and video editing detectors already achieve 82--97\% accuracy on such manipulations~\cite{rossler2019faceforensics,wang2020cnn,gragnaniello2021gan,croitoru2024deepfake,ni2026genvidbench}, making them a problem that complementary tools can address. The harder and more urgent challenge lies in detecting fully AI-generated videos: large-scale benchmarks such as GenVidBench~\cite{ni2026genvidbench} show that cross-generator detection accuracy for T2V and I2V content often falls to 42--70\%. It is precisely this gap that the Moir\'e motion invariant fills. No current generative model can synthesize the precise frame-to-frame coupling between fringe phase and grating assembly displacement that real optics produce, and splicing attacks are similarly defeated because the transplanted phase signal cannot match a different camera-assembly geometry. Our method thus complements existing forensic detectors in a layered verification pipeline: editing detectors screen for localized tampering, while the Moir\'e motion invariant provides a physics-grounded authentication layer against the generative attacks that remain beyond the reach of current detection methods.

\begin{figure}[t]
    \centering
    \includegraphics[width=0.99\linewidth]{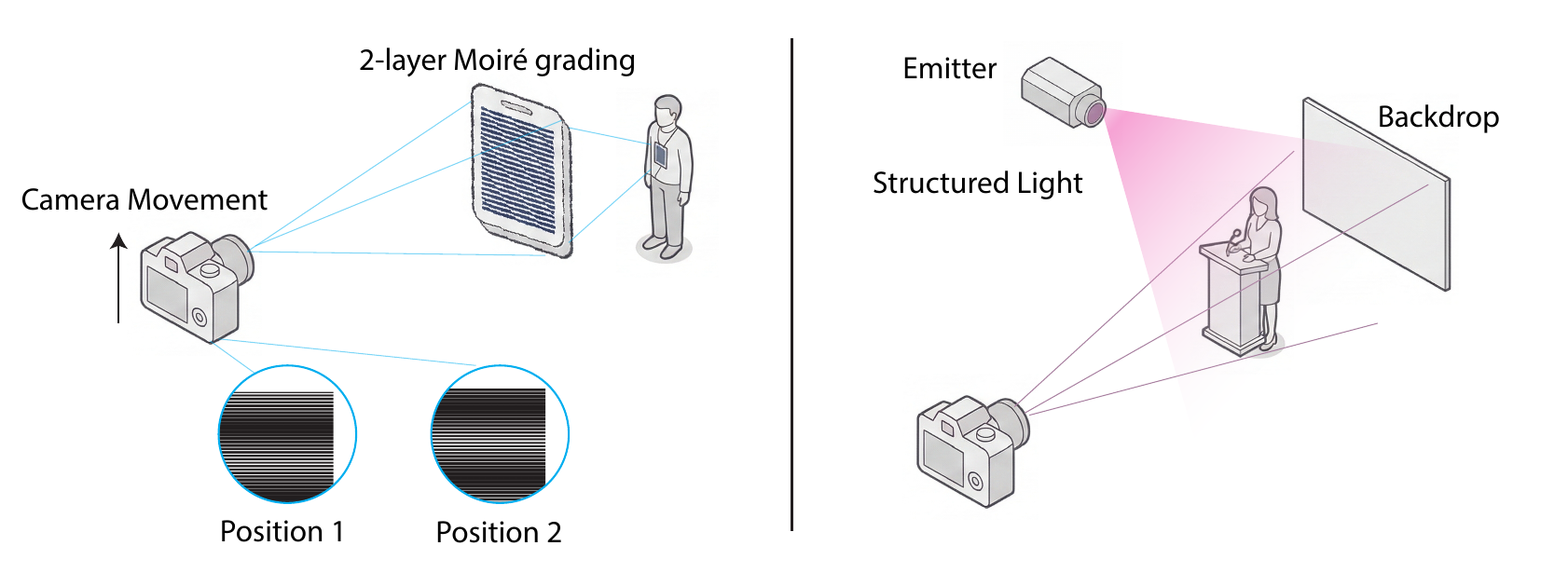}
    \caption{(Left) Our proposed Moiré-based signature requires only a passive 2-layer Moiré structure (e.g., worn as a badge), where standard camera movement naturally captures phase shifts. (Right) Active structured light signatures~\cite{schwartz2025verilight, michael2025noise}, while also using physical signals for authentication, require specialized external emitter hardware to project patterns onto the scene.}
    \label{fig:moire-vs-verilight}
\end{figure}
\section{Related Work}
\label{sec:related}

\paragraph{Moir\'e-based sensing and applications.}
Moir\'e patterns have recently been used as passive visual amplifiers for high-precision tracking. Moir\'eBoard~\cite{xiao2021moireboard} derives a calibration-free mapping from observed fringe geometry to 3-DoF camera position. Moir\'eTracker~\cite{moiretracker} characterizes screen-camera Moir\'e features for continuous 6-DoF camera-to-screen pose tracking. Moir\'eVision~\cite{moirevision} generalizes Moir\'e-based sensing to curvilinear patterns under practical perspective changes. These methods use Moir\'e patterns as measurement signals and derive explicit motion/pose recovery models. Our work differs in both application and derivation: rather than estimating pose, we eliminate the unknown camera-motion and distance terms to obtain a correlation-based invariant between fringe phase motion and grating image motion. The resulting test requires only that the two extracted video signals remain physically coupled, making Moir\'e motion a passive authentication signature instead of a tracking signal.

\paragraph{DeepFake and AI-generated video.} Our work is broadly motivated by the hyper-realistic generative media that calls for reliable systems to distinguish synthetic video from real footage; DeepFake~\cite{rossler2019faceforensics, croitoru2024deepfake} enables the convenient creation and dissemination of misinformation. Unlike authenticity techniques, afterthought defenses~\cite{Zheng_2025_ICCV, zhang2025physicsdriven, interno2025aigenerated, zhang2025factr1, kundu2025towards, li2025videohallu} are in an arms race with the rapid development of generative models~\cite{brooks2024sora,deepmind2025veo,kuaishou2024kling,esser2023structure,luma2024dream}.

\paragraph{Physics-based video forensics.} A separate line of work detects AI-generated video by identifying violations of physical plausibility, such as irregular eye pupil geometry~\cite{guo2022eyes}, inconsistent lighting and shadow directions~\cite{farid2022lighting}, or semantic and anatomical anomalies like extra fingers~\cite{tan2025semantic}. These methods exploit incidental physical errors that current generators happen to make. However, as generators improve, these artifacts are progressively eliminated through better training data and architectures. Our approach sidesteps this limitation: rather than searching for the absence of artifacts, we introduce a deliberate physical structure whose optical behavior is governed by deterministic laws that generators cannot learn to replicate statistically.

The work most closely related to ours are methods that  actively project temporally modulated light at the recording site to embed a verifiable physical signature into video, such as VeriLight~\cite{schwartz2025verilight} and Michael et al.~\cite{michael2025noise}. Our approach is similar in spirit, grounding authentication in physical-world signals, but differs fundamentally in that the Moir\'e signature is entirely passive: it arises naturally from the optical geometry of a compact grating structure and ordinary camera motion, requiring no active illumination or dedicated infrastructure (\cref{fig:moire-vs-verilight}). A complementary line of defense involves digital watermarking~\cite{hu2025videoshield, zhang2019RivaGan, zou2022antineuron}, which embeds provenance information at the software level.

\section{Conclusion and Limitation}
\label{sec:conclusion}

\paragraph{Limitation.}
Our approach relies on the assumption that current generative models do not perform physics-based simulation of Moir\'e optics. Should future models incorporate full ray-tracing of the two-layer grating geometry, they could in principle reproduce the correct fringe-displacement coupling. Additionally, the method requires relative motion between the camera and the grating assembly to generate a measurable phase signal; authentication is not possible in the case where both remain static throughout the recording.

To conclude, we presented a Moiré-based video authentication method that leverages a physics-governed, deterministic link between fringe phase evolution and camera-grating motion. We derived a Moiré motion invariant that is independent of viewing distance and camera intrinsics, and built a proof-of-concept system that extracts and tests it from video alone. Experiments show significantly different correlation values for real versus AI-generated videos, enabling verifiable discrimination between authentic and synthetic recordings.

Beyond the specific Moir\'e-based implementation, our work points to a broader principle: physical phenomena governed by deterministic optical laws can serve as authentication signatures that are fundamentally difficult for statistical generative models to forge. We hope this perspective inspires future exploration of other physics-based signatures for media authentication, complementing existing forensic and watermarking approaches to build more robust defenses against AI-generated misinformation.

\section*{Acknowledgements}
This work was supported in part by NSF 2540851 and a Gemini Academic Program Award. We thank Wenqi Wang, Yuwen Tan, Shengao Wang, Max Whitton, Michael Wakeham, Arjun Chandra, Joey Huang, Kevin Miller, and Audrey Mao for their help in testing our work in real recorded videos.


\bibliographystyle{splncs04}
\bibliography{main}

@String(CVPR  = {IEEE Conf. Comput. Vis. Pattern Recog.})

@String(ICCV  = {Int. Conf. Comput. Vis.})

@String(ECCV  = {Eur. Conf. Comput. Vis.})

@String(ICLR  = {Int. Conf. Learn. Represent.})

@String(AAAI  = {AAAI})

@String(ICASSP=	{ICASSP})

@String(ICME  = {Int. Conf. Multimedia and Expo})

@String(TOG   = {ACM Trans. Graph.})

@String(CVPR  = {CVPR})

@String(ICCV  = {ICCV})

@String(ECCV  = {ECCV})

@String(ICLR  = {ICLR})

@String(ICME  =	{ICME})

@String(TOG   = {ACM TOG})

@book{amidror2009moire,
  author    = {Amidror, Isaac},
  title     = {The Theory of the Moir{\'e} Phenomenon, Volume {I}: Periodic Layers},
  edition   = {2nd},
  publisher = {Springer},
  year      = {2009}
}

@inproceedings{xiao2021moireboard,
  title={Moir{\'e}board: A stable, accurate and low-cost camera tracking method},
  author={Xiao, Chang and Zheng, Changxi},
  booktitle={The 34th Annual ACM Symposium on User Interface Software and Technology},
  pages={881--893},
  year={2021}
}

@article{gabrielyan2007moire,
  author    = {Gabrielyan, Emin},
  title     = {The Basics of Line Moir{\'e} Patterns and Optical Speedup},
  journal   = {arXiv preprint arXiv:physics/0703098},
  year      = {2007}
}

@inproceedings{campos2024moirewidgets,
  title={{Moir{\'e}Widgets}: High-Precision, Passive Tangible Interfaces via Moir{\'e} Effect},
  author={Campos Zamora, Daniel and Dogan, Mustafa Doga and Siu, Alexa F and Koh, Eunyee and Xiao, Chang},
  booktitle={Proceedings of the 2024 CHI Conference on Human Factors in Computing Systems},
  pages={1--10},
  year={2024}
}

@misc{chen2024deepfake,
  author    = {CNN},
  title     = {Finance worker pays out \$25 million after video call with deepfake chief financial officer},
  year      = {2024},
}

@misc{brooks2024sora,
  author    = {Brooks, Tim and Peebles, Bill and Holmes, Connor and DePue, Will and Guo, Yufei and Jing, Li and Schnurr, David and Taylor, Joe and Luhman, Troy and Luhman, Eric and Ng, Clarence and Wang, Ricky and Ramesh, Aditya},
  title     = {Video Generation Models as World Simulators},
  howpublished = {OpenAI Technical Report},
  year      = {2024},
}

@misc{runway2025gen45,
  author    = {{Runway}},
  title     = {Introducing {Runway Gen-4.5}},
  howpublished = {Runway Research},
  year      = {2025},
}

@misc{bytedance2026seedance,
  author    = {{ByteDance Seed Team}},
  title     = {Seedance 2.0},
  howpublished = {ByteDance Seed},
  year      = {2026},
}

@inproceedings{wang2020cnn,
  author    = {Wang, Sheng-Yu and Wang, Oliver and Zhang, Richard and Owens, Andrew and Efros, Alexei A.},
  title     = {{CNN}-Generated Images Are Surprisingly Easy to Spot\ldots for Now},
  booktitle = CVPR,
  year      = {2020}
}

@inproceedings{gragnaniello2021gan,
  author    = {Gragnaniello, Diego and Cozzolino, Davide and Marra, Francesco and Poggi, Giovanni and Verdoliva, Luisa},
  title     = {Are {GAN} Generated Images Easy to Detect? {A} Critical Analysis of the State-of-the-Art},
  booktitle = ICME,
  year      = {2021}
}

@misc{c2pa2022spec,
  author    = {{Coalition for Content Provenance and Authenticity (C2PA)}},
  title     = {{C2PA} Technical Specification},
  howpublished = {C2PA},
  year      = {2022},
}

@misc{deepmind2023synthid,
  author    = {{Google DeepMind}},
  title     = {{SynthID}: Identifying {AI}-Generated Content},
  howpublished = {Google DeepMind},
  year      = {2023},
}

@article{croitoru2024deepfake,
  author    = {Croitoru, Florinel-Alin and Hiji, Andrei-Iulian and Hondru, Vlad and Ristea, Nicolae-Catalin and Irofti, Paul and Popescu, Marius and Rusu, Cristian and Ionescu, Radu Tudor and Khan, Fahad Shahbaz and Shah, Mubarak},
  title     = {Deepfake Media Generation and Detection in the Generative {AI} Era: A Survey and Outlook},
  journal   = {arXiv preprint arXiv:2411.19537},
  year      = {2024}
}

@misc{liggett2025deepfake,
  author    = {BBC},
  title     = {Disgraceful Deep-fake {AI} video condemned by presidential candidate},
  year      = {2025},
}

@article{takasaki1970moire,
  author    = {Takasaki, Hiroshi},
  title     = {Moir{\'e} Topography},
  journal   = {Applied Optics},
  volume    = {9},
  number    = {6},
  pages     = {1467--1472},
  year      = {1970}
}

@article{sciammarella1982moire,
  author    = {Sciammarella, Cesar A.},
  title     = {The Moir{\'e} Method -- A Review},
  journal   = {Experimental Mechanics},
  volume    = {22},
  number    = {11},
  pages     = {418--433},
  year      = {1982}
}

@book{kafri1990moire,
  author    = {Kafri, Oded and Glatt, Ilana},
  title     = {The Physics of Moir{\'e} Metrology},
  publisher = {Wiley},
  year      = {1990}
}

@article{hersch2004moire,
  author    = {Hersch, Roger D. and Chosson, S{\'e}bastien},
  title     = {Band Moir{\'e} Images},
  journal   = TOG,
  volume    = {23},
  number    = {3},
  pages     = {239--248},
  year      = {2004}
}

@inproceedings{sethapakdi2025fabobscura,
  author    = {Sethapakdi, Theophanis and Perroni-Scharf, Matteo and Li, Meng and Li, Jingyi and Solomon, Justin and Satyanarayan, Arvind and Mueller, Stefanie},
  title     = {{FabObscura}: Computational Design and Fabrication for Interactive Barrier-Grid Animations},
  booktitle = {Proceedings of the ACM Symposium on User Interface Software and Technology (UIST)},
  year      = {2025}
}

@article{zhang2000flexible,
  title={A flexible new technique for camera calibration},
  author={Zhang, Zhengyou},
  journal={IEEE Transactions on pattern analysis and machine intelligence},
  volume={22},
  number={11},
  pages={1330--1334},
  year={2000},
  publisher={Ieee}
}

@article{liu2023deepfacelab,
  title={Deepfacelab: Integrated, flexible and extensible face-swapping framework},
  author={Liu, Kunlin and Perov, Ivan and Gao, Daiheng and Chervoniy, Nikolay and Zhou, Wenbo and Zhang, Weiming},
  journal={Pattern Recognition},
  volume={141},
  pages={109628},
  year={2023},
  publisher={Elsevier}
}

@misc{facefusion2024,
  author    = {{FaceFusion}},
  title     = {{FaceFusion}: Industry Leading Face Manipulation Platform},
  howpublished = {GitHub},
  year      = {2024}
}

@incollection{zuiderveld1994clahe,
  author    = {Zuiderveld, Karel},
  title     = {Contrast Limited Adaptive Histogram Equalization},
  booktitle = {Graphics Gems {IV}},
  publisher = {Academic Press},
  pages     = {474--485},
  year      = {1994}
}

@article{suzuki1985contours,
  author    = {Suzuki, Satoshi and Abe, Keiichi},
  title     = {Topological Structural Analysis of Digitized Binary Images by Border Following},
  journal   = {Computer Vision, Graphics, and Image Processing},
  volume    = {30},
  number    = {1},
  pages     = {32--46},
  year      = {1985}
}

@misc{blender2025,
  author       = {{Blender Foundation}},
  title        = {Blender: Open Source 3D Creation Suite},
  year         = {2025},
  organization = {Blender Foundation},
  url          = {https://www.blender.org}
}

@techreport{bouguet2001lucas,
  author    = {Bouguet, Jean-Yves},
  title     = {Pyramidal Implementation of the {Lucas Kanade} Feature Tracker: Description of the Algorithm},
  institution = {Intel Corporation, Microprocessor Research Labs},
  year      = {2001}
}

@inproceedings{ni2026genvidbench,
  author    = {Ni, Zhisheng and Yan, Qiuyu and Huang, Mengqi and Yuan, Tao and Tang, Yifan and Hu, Haiyang and Chen, Xin and Wang, Yaowei},
  title     = {{GenVidBench}: A 6-Million Benchmark for {AI}-Generated Video Detection},
  booktitle = AAAI,
  year      = {2026}
}

@misc{deepmind2025veo,
  author       = {{Google DeepMind}},
  title        = {Veo 3.1},
  year         = {2025},
  organization = {Google DeepMind},
  url          = {https://deepmind.google/}
}

@misc{xai2025grok,
  author       = {{xAI}},
  title        = {Grok Imagine Video},
  year         = {2025},
  organization = {xAI},
  url          = {https://x.ai}
}

@article{hacohen2026ltx2,
  author    = {HaCohen, Yoav and Brazowski, Bar and Chiprut, Nisan and Bitterman, Yaron and Kvochko, Anna and Berkowitz, Alon and Shalem, Dor and Lifschitz, Daniel and Moshe, Dana and Porat, Eyal and Richardson, Elad and Shiran, Guy and Chachy, Itai and Chetboun, Jonathan and Finkelson, Matan and Kupchick, Michael and Zabari, Nir and Guetta, Nadav and Kotler, Niv and Bibi, Ofir and Gordon, Ori and Panet, Philippe and Benita, Rami and Armon, Shachar and Kulikov, Vladimir and Inger, Yaki and Shiftan, Yonadav and Melumian, Zeev and Farbman, Zeev},
  title     = {{LTX-2}: Efficient Joint Audio-Visual Foundation Model},
  journal   = {arXiv preprint arXiv:2601.03233},
  year      = {2026}
}

@inproceedings{schwartz2025verilight,
  author    = {Schwartz, Hadleigh and Yan, Xiaofeng and Carver, Charles J. and Zhou, Xia},
  title     = {Combating Falsification of Speech Videos with Live Optical Signatures},
  booktitle = {Proceedings of the 2025 ACM SIGSAC Conference on Computer and Communications Security (CCS)},
  year      = {2025},
  pages     = {3296--3310},
  doi       = {10.1145/3719027.3765112}
}

@inproceedings{hu2025videoshield,
  title={VideoShield: Regulating Diffusion-based Video Generation Models via Watermarking},
  author={Hu, Runyi and Zhang, Jie and Li, Yiming and Li, Jiwei and Guo, Qing and Qiu, Han and Zhang, Tianwei},
  booktitle={International Conference on Learning Representations (ICLR)},
  year={2025},
}

@article{zhang2019RivaGan,
  title={Robust Invisible Video Watermarking with Attention},
  author={Zhang, Kevin Alex and Xu, Lei and Cuesta-Infante, Alfredo and Veeramachaneni, Kalyan},
  journal={arXiv preprint arXiv:1909.01285},
  year={2019},
}

@inproceedings{zou2022antineuron,
  title={Anti-Neuron Watermarking: Protecting Personal Data Against Unauthorized Neural Networks},
  author={Zou, Zihang and Gong, Boqing and Wang, Liqiang},
  booktitle={European Conference on Computer Vision (ECCV)},
  pages={449--465},
  year={2022},
  organization={Springer}
}

@inproceedings{interno2025aigenerated,
  title={AI-Generated Video Detection via Perceptual Straightening},
  author={Intern{\`o}, Christian and Geirhos, Robert and Olhofer, Markus and Liu, Sunny and Hammer, Barbara and Klindt, David},
  booktitle={Advances in Neural Information Processing Systems},
  volume={38},
  year={2025}
}

@InProceedings{Zheng_2025_ICCV,
    author    = {Zheng, Chende and Suo, Ruiqi and Lin, Chenhao and Zhao, Zhengyu and Yang, Le and Liu, Shuai and Yang, Minghui and Wang, Cong and Shen, Chao},
    title     = {D3: Training-Free AI-Generated Video Detection Using Second-Order Features},
    booktitle = {Proceedings of the IEEE/CVF International Conference on Computer Vision (ICCV)},
    month     = {October},
    year      = {2025},
    pages     = {12852-12862}
}

@inproceedings{zhang2025factr1,
  title={{Fact-R1}: Towards Explainable Video Misinformation Detection with Deep Reasoning},
  author={Zhang, Fanrui and Li, Dian and Zhang, Qiang and Chen, Jun and Liu, Gang and Lin, Junxiong and Yan, Jiahong and Liu, Jiawei and Zha, Zheng-Jun},
  booktitle={Advances in Neural Information Processing Systems},
  volume={38},
  year={2025}
}

@inproceedings{zhang2025physicsdriven,
  title={Physics-Driven Spatiotemporal Modeling for AI-Generated Video Detection},
  author={Zhang, Shuhai and Lian, Zihao and Yang, Jiahao and Li, Daiyuan and Pang, Guoxuan and Liu, Feng and Han, Bo and Li, Shutao and Tan, Mingkui},
  booktitle={Advances in Neural Information Processing Systems},
  volume={38},
  year={2025}
}

@inproceedings{kundu2025towards,
  title={Towards a Universal Synthetic Video Detector: From Face or Background Manipulations to Fully AI-Generated Content},
  author={Kundu, Rohit and Xiong, Hao and Mohanty, Vishal and Balachandran, Athula and Roy-Chowdhury, Amit K.},
  booktitle={Proceedings of the IEEE/CVF Conference on Computer Vision and Pattern Recognition (CVPR)},
  year={2025},
  publisher={IEEE}
}

@article{michael2025noise,
  author    = {Michael, Peter F. and Hao, Zekun and Belongie, Serge and Davis, Abe},
  title     = {Noise-Coded Illumination for Forensic and Photometric Video Analysis},
  journal   = {ACM Transactions on Graphics},
  year      = {2025},
  volume    = {44},
  number    = {5},
  articleno = {165},
  numpages  = {16},
  doi       = {10.1145/3742892}
}

@inproceedings{li2025videohallu,
  title={VideoHallu: Evaluating and Mitigating Multi-modal Hallucinations on Synthetic Video Understanding},
  author={Li, Zongxia and Wu, Xiyang and Shi, Guangyao and Qin, Yubin and Du, Hongyang and Zhou, Tianyi and Manocha, Dinesh and Boyd-Graber, Jordan Lee},
  booktitle={Advances in Neural Information Processing Systems},
  volume={38},
  year={2025}
}

@inproceedings{rossler2019faceforensics,
  author = {Rossler, Andreas and Cozzolino, Davide and Verdoliva, Luisa and Riess, Christian and Thies, Justus and Niessner, Matthias},
title = {FaceForensics++: Learning to Detect Manipulated Facial Images},
booktitle = {Proceedings of the IEEE/CVF International Conference on Computer Vision (ICCV)},
month = {October},
year = {2019}
}

@misc{kuaishou2024kling,
  title={Kling: A pioneering AI video generation model},
  author={{Kuaishou Technology}},
  howpublished={\url{https://kling.kuaishou.com/}},
  year={2024}
}

@inproceedings{esser2023structure,
  title={Structure and Content-Guided Video Synthesis with Diffusion Models},
  author={Esser, Patrick and Chiu, Johnathan and Atighehchian, Parmida and Granskog, Jonathan and Germanidis, Anastasios},
  booktitle={Proceedings of the IEEE/CVF International Conference on Computer Vision (ICCV)},
  pages={7346--7356},
  year={2023},
  publisher={IEEE}
}

@misc{luma2024dream,
  title={Dream Machine: High quality, realistic video generation from text and images},
  author={{Luma AI}},
  howpublished={\url{https://lumalabs.ai/dream-machine}},
  year={2024}
}

@inproceedings{guo2022eyes,
  author    = {Guo, Hui and Hu, Shu and Wang, Xin and Chang, Ming-Ching and Lyu, Siwei},
  title     = {Eyes Tell All: Irregular Pupil Shapes Reveal {GAN}-Generated Faces},
  booktitle = {IEEE International Conference on Acoustics, Speech and Signal Processing (ICASSP)},
  year      = {2022},
  pages     = {2904--2908},
  organization = {IEEE}
}

@article{farid2022lighting,
  title={Lighting (in) consistency of paint by text},
  author={Farid, Hany},
  journal={arXiv preprint arXiv:2207.13744},
  year={2022}
}

@article{tan2025semantic,
  title={Semantic Visual Anomaly Detection and Reasoning in AI-Generated Images},
  author={Tan, Chuangchuang and Ming, Xiang and Wang, Jinglu and Tao, Renshuai and Li, Bin and Wei, Yunchao and Zhao, Yao and Lu, Yan},
  journal={arXiv preprint arXiv:2510.10231},
  year={2025}
}

@ARTICLE{moiretracker,
  author={Ning, Jingyi and Xie, Lei and Li, Yi and Chen, Yingying and Bu, Yanling and Wang, Chuyu and Lu, Sanglu and Ye, Baoliu},
  journal={IEEE Journal on Selected Areas in Communications}, 
  title={MoiréTracker: Continuous Camera-to-Screen 6-DoF Pose Tracking Based on Moiré Pattern}, 
  year={2024},
  volume={42},
  number={10},
  pages={2642-2658},
  doi={10.1109/JSAC.2024.3414619}}

@inproceedings{moirevision,
author = {Ning, Jingyi and Xie, Lei and Yan, Zhihao and Bu, Yanling and Luo, Jun},
title = {Moir\'{e}Vision: A Generalized Moir\'{e}-based Mechanism for 6-DoF Motion Sensing},
year = {2024},
isbn = {9798400704895},
publisher = {Association for Computing Machinery},
address = {New York, NY, USA},
url = {https://doi.org/10.1145/3636534.3649374},
doi = {10.1145/3636534.3649374},
booktitle = {Proceedings of the 30th Annual International Conference on Mobile Computing and Networking},
pages = {467–481},
numpages = {15},
location = {Washington D.C., DC, USA},
series = {ACM MobiCom '24}
}


\clearpage
\appendix
\section*{\centering \Large Supplementary Material}
\renewcommand\thesection{\Alph{section}}
\renewcommand\thesubsection{\thesection.\arabic{subsection}}

This supplementary material provides additional details, experimental results, and discussion to support the main manuscript. It is organized as follows:

\begin{itemize}
    \item \cref{appendix:setup}: Detailed camera configurations, experiment setup, and instructions for reproducing our physical device.
    \item \cref{appendix:attack}: A detailed description of the attack process.
    \item \cref{appendix:results}: Additional experimental results, including ROC curves, threshold-based classification, and robustness tests under alternative pattern layouts and 3D slant.
    \item \cref{appendix:discussion}: Extended discussion and future work, addressing potential limitations and real-world scalability.
\end{itemize}

\section{Camera Configurations and Hardware Setup}
\label{appendix:setup}
Data were acquired using an iPhone 15 (1080p, 60 fps). To capture the Moiré effect, we employed dynamic camera movements (translation and varying viewing distance) relative to the Moiré board. This intentional motion is essential, as the captured interference patterns are inherently dependent on the perspective and relative displacement between the camera and the dual-grating layers. We utilized the camera’s auto-exposure mechanism to adapt to the varying luminance of the interference fringes throughout the recording, ensuring consistent signal visibility without manual exposure locking.

\subsection{Reproducing the Device}
\label{appendix:reproduction}
To facilitate the reproduction of the Moiré board in any laboratory environment, we provide the following hardware specifications and assembly instructions.

\subsubsection{Materials and Digital Assets}
The device consists of a three-layer optical stack. The primary components are as follows:
\begin{itemize}
    \item \textbf{Base Layer (Support):} A 3mm (1/8 inch) thick clear plexiglass sheet (8" $\times$ 10") provides structural support. \href{https://www.amazon.com/Plexiglass-Sheets-3mm-Acrylic-Engraver/dp/B0FLXNY6S8/ref=sr_1_1?crid=3N4CGCPF7WQ2M&dib=eyJ2IjoiMSJ9.XYQSJUjvNJsKB-SxDXZFgj6MfBTOk8W1Tbiu_v2CcY3Nk8OnJ0QoNpsFAgx_ofgPf_LXYrRvOG9O_FTRsNkQZGDGa4gByJCv8354wkKdW5Gv-YRTKiztBZHB0j--ZdPRzpbCkOqPKKZ1sIUk4udvaphXgjDIee8K6_eRun-CrH3GUyz-IettCu8riPBioqwsaFpAfjje5N4clsKK9-0Xn-2F-j5MGJJlLsaA7zP7aT8.R7WEbHvwRYURSenEkPcuKeH1qpB6ZxTbo3FvUT0ub6s&dib_tag=se&keywords=(10%2BPack)%2B8%22%2Bx%2B10%22%2BPlexiglass%2BSheets%2B1%2F8%2Binch%2BThick%2B(3mm)%2C%2BClear%2BAcrylic%2BSheets%2BPanel%2BCut%2Bto%2BSize%2BPlexiglass%2BAcrylic%2BBoard%2C%2BCut%2Bwith%2BEngraver%2C%2BPower%2BSaw%2Bor%2BHand%2BTools%2C%2BNo%2BLaser%2BCut.&nsdOptOutParam=true&qid=1773246105&sprefix=10%2Bpack%2B8%2Bx%2B10%2Bplexiglass%2Bsheets%2B1%2F8%2Binch%2Bthick%2B3mm%2B%2C%2Bclear%2Bacrylic%2Bsheets%2Bpanel%2Bcut%2Bto%2Bsize%2Bplexiglass%2Bacrylic%2Bboard%2C%2Bcut%2Bwith%2Bengraver%2C%2Bpower%2Bsaw%2Bor%2Bhand%2Btools%2C%2Bno%2Blaser%2Bcut.%2Caps%2C156&sr=8-1&th=1}{[Amazon Link]}
    \item \textbf{Rear Layer (Reference Grating):} A printed pattern consisting of 160 horizontal lines with a pitch of $p_r = 0.52$ mm.
    \item \textbf{Front Layer (Viewing Grating):} A 1mm-thick lenticular lens sheet (50 LPI, $p_f = 0.508$ mm). \href{https://www.amazon.com/Lenticular-Adhesive-Motion-Printing-Sheets/dp/B0DTYZ9KPH/ref=sr_1_1?crid=109HSELNGYCVH&dib=eyJ2IjoiMSJ9.rZ5aITG8DyRVpqRXu0IeK-mMDcWkJ68UU4rgvzLXny2YGwL0wII0bIqNlW3t91sk_aV_Hnc36X_e0v1UfvmxRSTW8KDho6b50wuTNBQuvuBy66ICWT0kdZFxai1ycm_OzwdFrJY5YAd-p-vQ9xrsa1_BRmZZ9toLcWAImB2uB8oNUJJj90rmteOehAF4N6Cb1kGg11ixoPnVef_HlwvxiQ.Bxoz4sikmtn69Kq6hG5Doaik1JPmZADue896zjPaKpI&dib_tag=se&keywords=Lenticular%2BSheet%2B50%2BLPI%2B4%22x6%22%2Bwith%2BAdhesive%2BBack%2Bfor%2BFlip%2C%2BMotion%2C%2B3D%2BPhoto%2BPrinting%2B(Pack%2Bof%2B10%2BSheets)%2B(Brand%2B%3A%2Bprint6D)&nsdOptOutParam=true&qid=1773246152&sprefix=lenticular%2Bsheet%2B50%2Blpi%2B4%2Bx6%2Bwith%2Badhesive%2Bback%2Bfor%2Bflip%2C%2Bmotion%2C%2B3d%2Bphoto%2Bprinting%2Bpack%2Bof%2B10%2Bsheets%2Bbrand%2Bprint6d%2B%2Caps%2C103&sr=8-1&th=1}{[Amazon Link]}
    \item \textbf{Digital Assets:} The reference grating is provided as a printable PDF (\texttt{reference grating.pdf}) in the root directory of the attached supplementary folder. To maintain the designed geometry (pitch $p_r = 0.52$ mm), the document must be printed at 100\% scale (actual size) on A4 paper.
\end{itemize}

\subsubsection{Assembly Instructions}
The assembly requires careful layer alignment:
\begin{enumerate}
    \item Place the printed Rear Layer (A4 paper) onto the 3mm Base Layer.
    \item Overlay the Front Layer (Lenticular Sheet) ensuring the grating lines of both layers are parallel.
    \item Secure the layers together to minimize the air gap between the two gratings, as this gap is critical for maintaining high-contrast Moiré interference across varying viewing angles and distances.
\end{enumerate}

\section{Detailed Description of the Attack Process}
\label{appendix:attack}

To comprehensively evaluate the robustness of our proposed Moiré video authentication method against state-of-the-art AI generation, we meticulously designed an adversarial video synthesis pipeline. This section details our model selection, prompt engineering strategy, dataset curation statistics, and the full prompt templates used to synthesize the fake videos.

\subsection{Model Selection}
We selected three state-of-the-art video generation models for both Text-to-Video (T2V) and Image-to-Video (I2V) capabilities. To ensure a comprehensive evaluation across different model architectures and availability levels, we chose:
\begin{itemize}
    \item \textbf{Veo 3.1 Pro~\cite{deepmind2025veo}}: Representing the industry-leading, closed-source commercial models with high fidelity and temporal consistency.
    \item \textbf{Grok Imagine Video~\cite{xai2025grok}}: Representing a highly competitive, alternative closed-source generative architecture.
    \item \textbf{LTX-2 Video~\cite{hacohen2026ltx2}}: Representing the current state-of-the-art in open-source video generation models, allowing us to evaluate publicly accessible attack vectors.
\end{itemize}

\subsection{Adversarial Video Synthesis and Dataset Curation}
\label{appendix:synthesis_stats}
To synthesize the adversarial attack videos, we targeted three distinct kinematic paradigms: (1) static subject with a moving camera, (2) moving subject with a static camera, and (3) simultaneous movement of both subject and camera.

\vspace{1ex}
\noindent\textbf{Controlled Prompt Strategy.} Designing prompts for video generation is inherently subjective. Rather than heavily engineering idiosyncratic prompts tailored to the specific inductive biases of each individual model, we deliberately formulated a unified, rigorous set of prompt templates applied uniformly across all three models. This standardizes the evaluation baseline, acting as a controlled zero-shot test of each model's intrinsic ability to adhere to complex physical and geometric constraints (i.e., maintaining the Moiré board's spatial integrity and phase structures) without model-specific prompt overfitting.

\vspace{1ex}
\noindent\textbf{Generation Yield and Stringent Filtering.}
To construct a robust evaluation foundation, we first collected a dataset comprising 87 authentic Moiré videos captured across both indoor (75 videos) and outdoor (12 videos) environments.

\begin{figure}[htbp]
\centering
\includegraphics[width=0.32\linewidth]{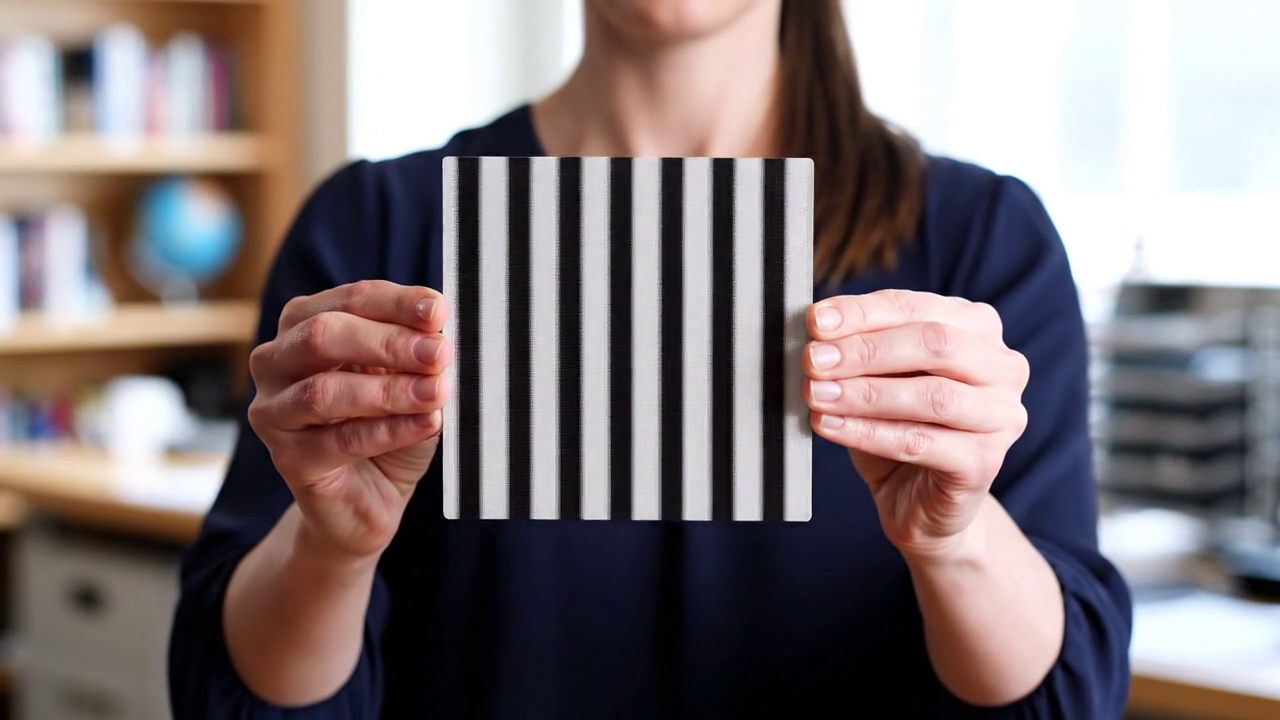}
\includegraphics[width=0.32\linewidth]{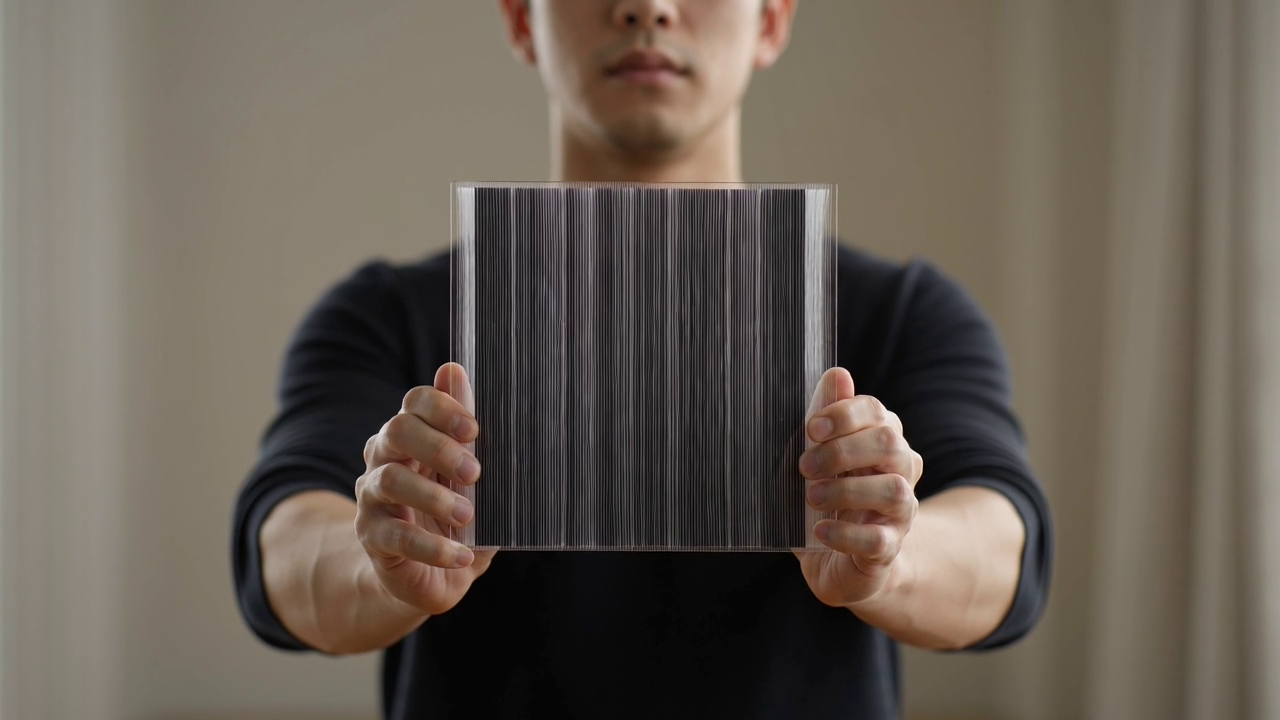}
\includegraphics[width=0.32\linewidth]{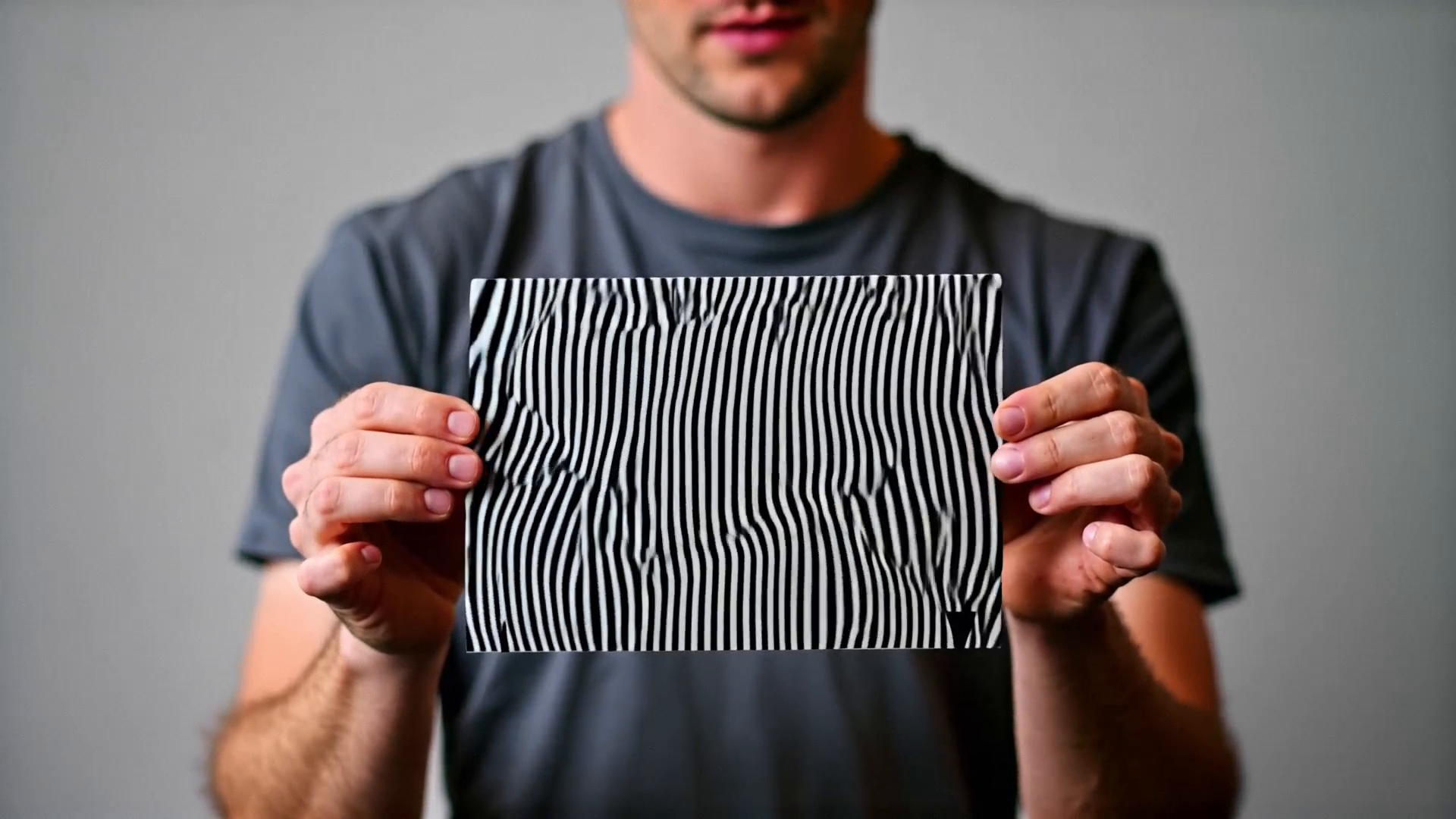}
\caption{Examples of pure Text-to-Video (T2V) generation failures. The models struggle to synthesize accurate, rigid Moiré patterns from scratch, frequently resulting in unnaturally thick stripes or severely wobbly, non-physical deformations.}
\label{fig:t2v_examples}
\end{figure}

Using the three selected state-of-the-art models, we initially explored pure Text-to-Video (T2V) generation. For each motion paradigm, we meticulously engineered specific text prompts and refined them over 15 to 20 iterations. Using the best-performing prompts, we generated 20 videos per model. However, empirical observations revealed that these 60 purely text-generated videos lacked physical realism and could be easily distinguished by the naked eye (see \cref{fig:t2v_examples} for representative artifacts).

Consequently, we shifted our focus to a first-frame-conditioned Image-to-Video (I2V) strategy. To significantly enhance visual fidelity and physical consistency, we extracted the initial frame from the authentic videos to serve as a visual anchor. We then randomly assigned an optimized motion prompt to guide the synthesis. This approach yielded 257 I2V generated videos, culminating in an initial synthetic pool of 317 videos (60 T2V and 257 I2V videos).

\begin{figure}[htbp]
\centering
\includegraphics[width=0.48\linewidth]{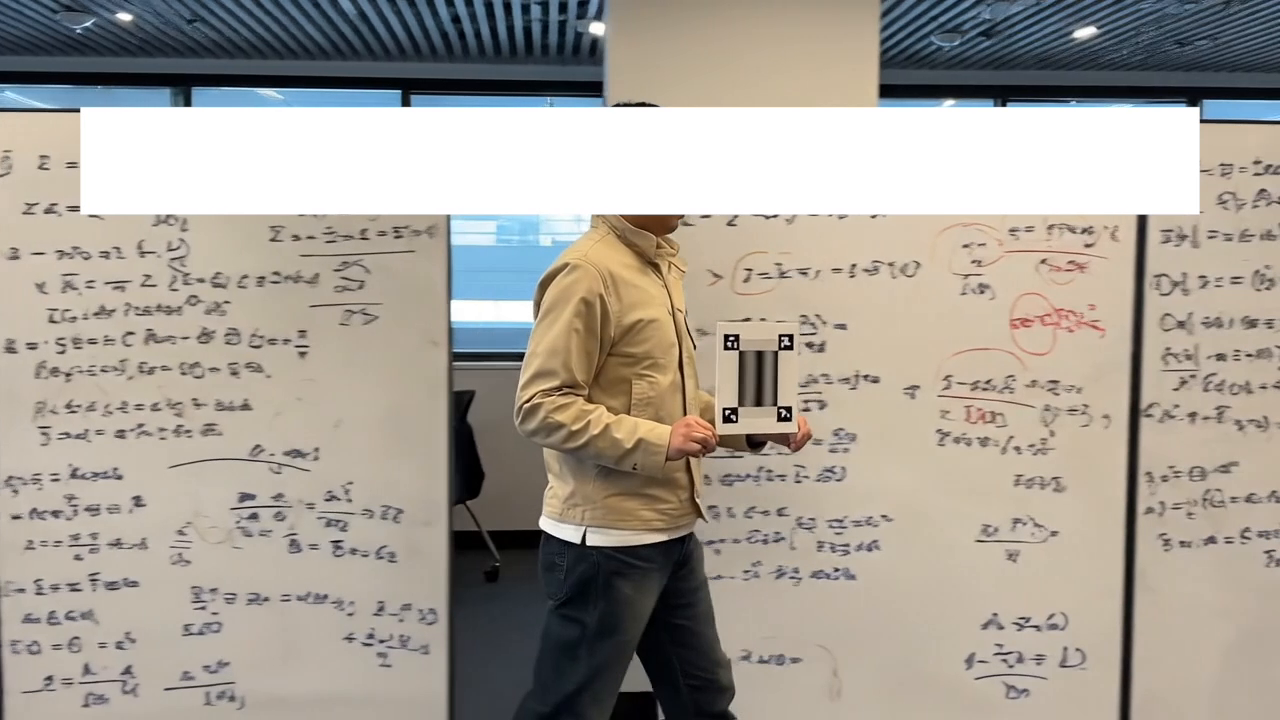}
\includegraphics[width=0.48\linewidth]{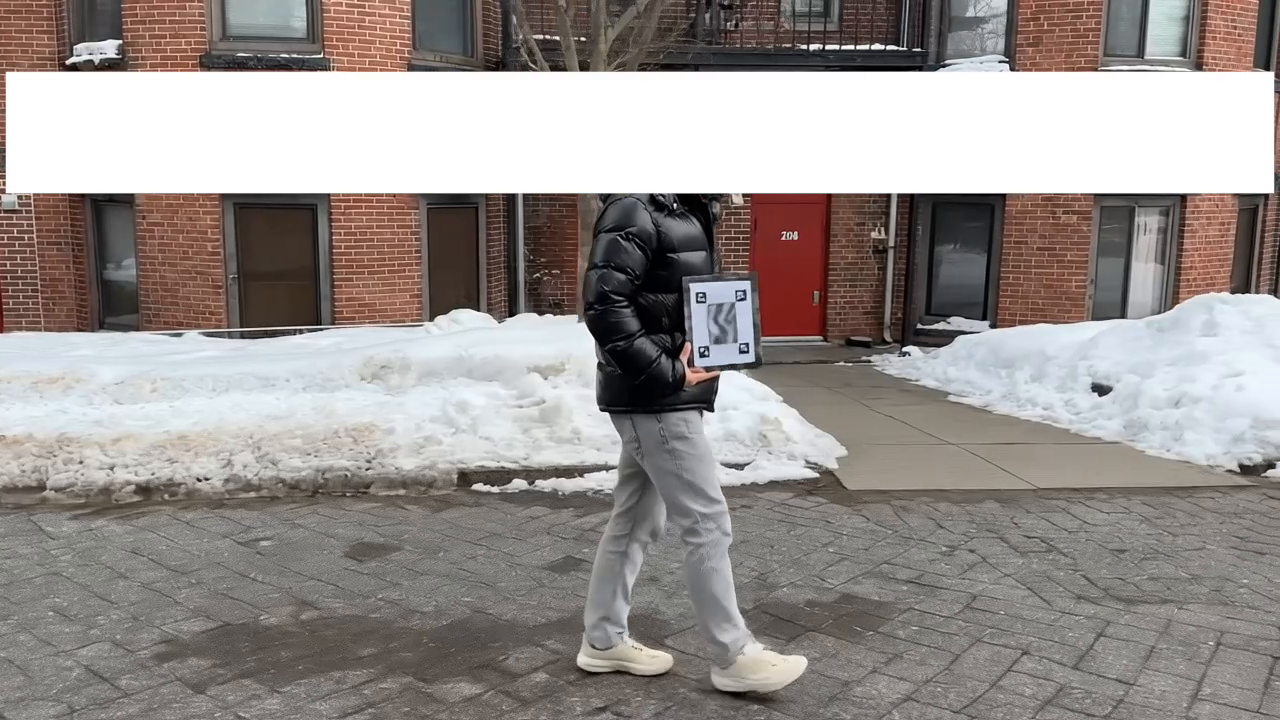} \
\vspace{1ex}
\includegraphics[width=0.32\linewidth]{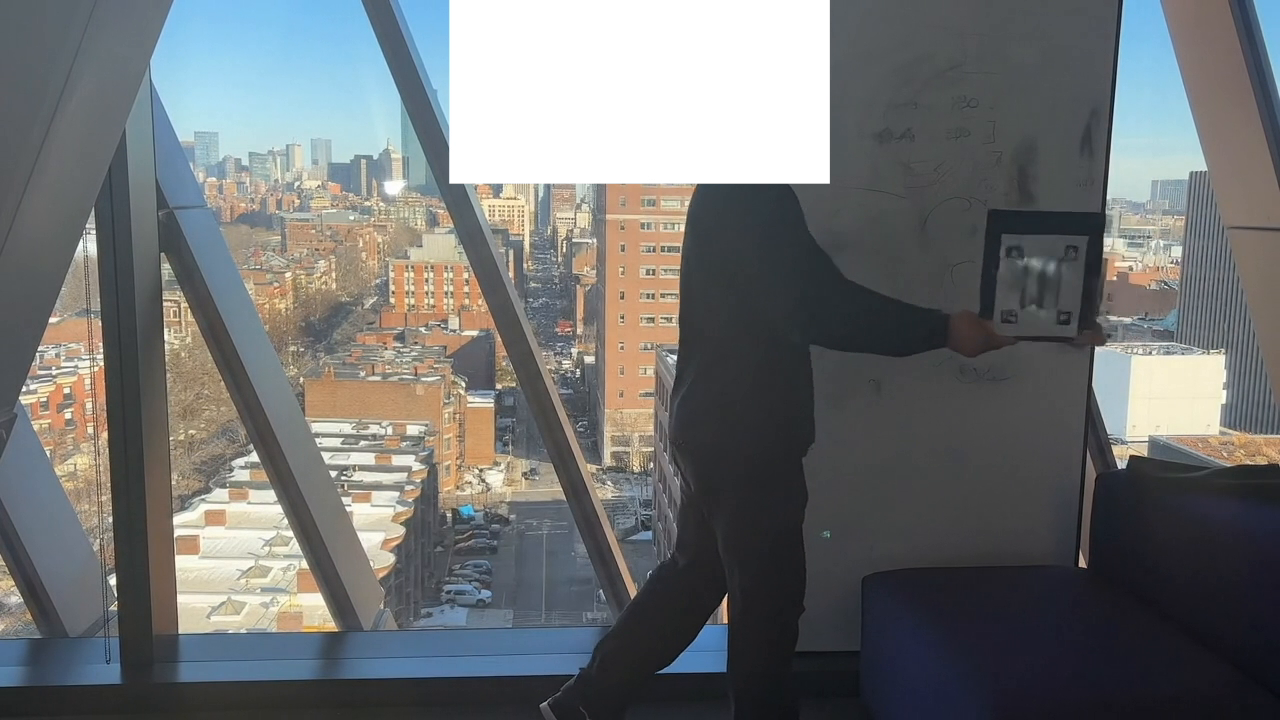}
\includegraphics[width=0.32\linewidth]{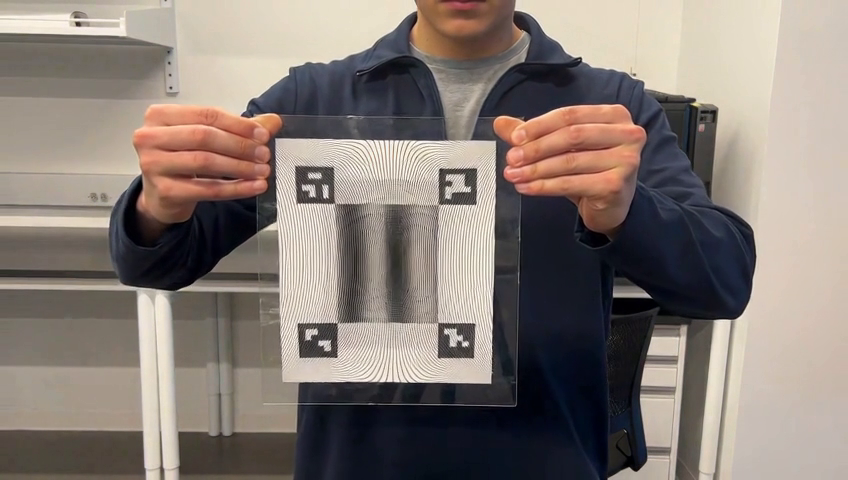}
\includegraphics[width=0.32\linewidth]{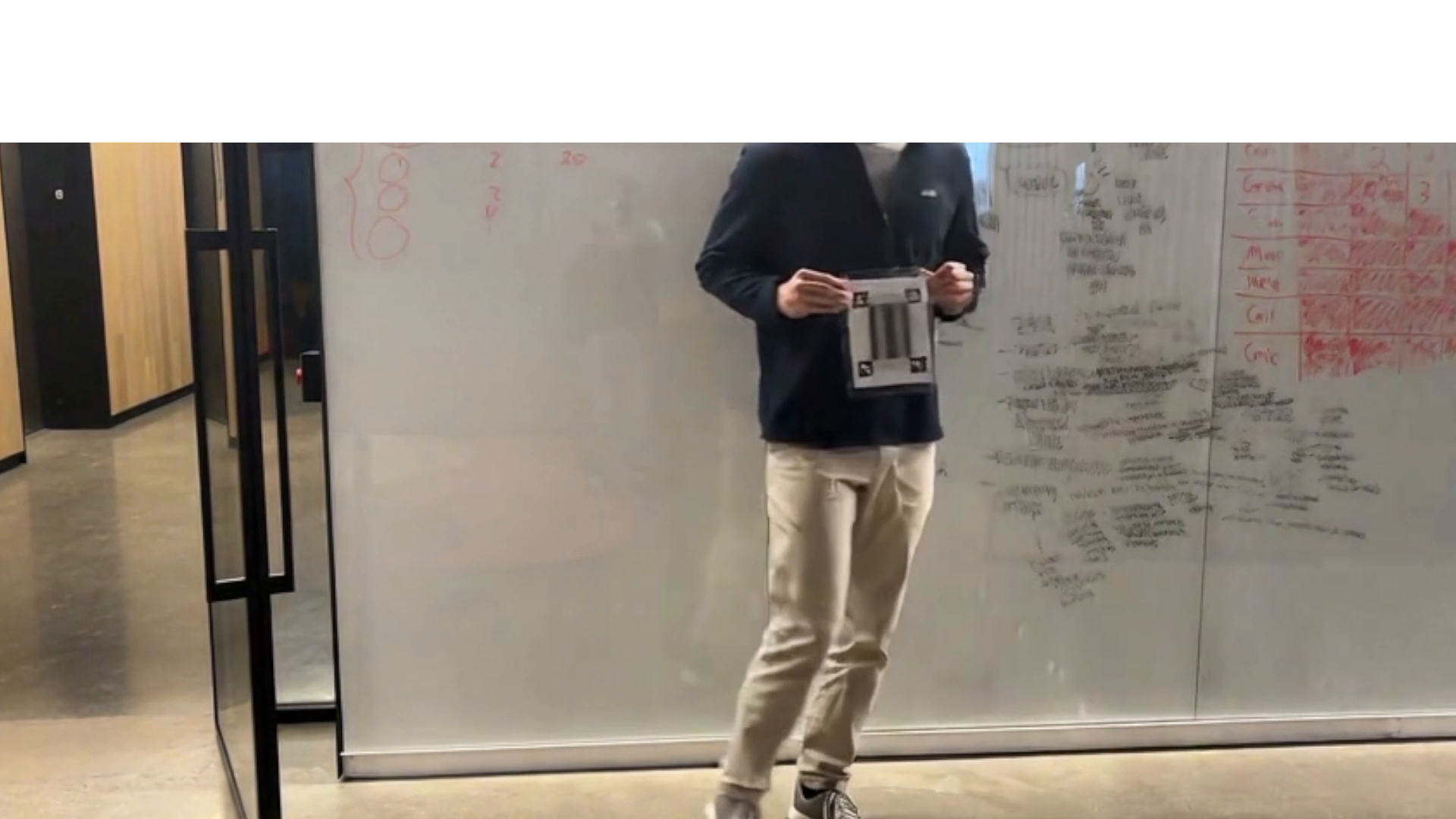}
\caption{Representative frames from Image-to-Video (I2V) generation exhibiting temporal and physical inconsistencies. While first-frame conditioning significantly improves overall fidelity, models still struggle with maintaining the rigid geometry of the grating during motion, resulting in structural warping, blurring, and detachment from the subject's hands.}
\label{fig:i2v_artifacts}
\end{figure}

\vspace{1ex}
\noindent\textbf{Final Curated Dataset for Evaluation.}
As discussed in the main manuscript, manual inspection revealed that generative models frequently produced severe temporal inconsistencies, such as the sudden disappearance, detachment, or extreme deformation of the Moiré grating assembly (\cref{fig:i2v_artifacts} illustrates examples of these persistent artifacts).

We screened and discarded 225 such corrupted samples. This curation yielded a final dataset of \textbf{92 high-quality synthetic videos} that appear deceptively authentic to the human eye. \textbf{All subsequent algorithmic evaluations, threshold analyses, and ROC metric computations were conducted exclusively on this challenging 92-video subset} alongside the 87 authentic videos.

\cref{tab:dataset_pipeline} summarizes the dataset generation and curation pipeline, detailing the exact yield per generative model. Additionally, \cref{tab:dataset_metadata} provides a comprehensive statistical breakdown of the video metadata for the final evaluation dataset, reflecting the native generation formats of the respective models.

\begin{table}[htbp]
\centering
\caption{Dataset Generation and Curation Pipeline. The initial synthetic pool of 317 videos was rigorously filtered to remove 225 samples with severe temporal or physical collapses, yielding 92 high-quality adversarial videos.}
\label{tab:dataset_pipeline}
\footnotesize
\begin{tabular}{lcccc}
\toprule
\textbf{Pipeline Stage} & \textbf{Veo 3.1 Pro} & \textbf{Grok Imagine} & \textbf{LTX-2 Video} & \textbf{Total} \\
\midrule
T2V Generated & 20 & 20 & 20 & 60 \\
I2V Generated (First-Frame) & 83 & 87 & 87 & 257 \\
\midrule
\textbf{Initial Synthetic Pool} & 103 & 107 & 107 & \textbf{317} \\
Discarded (Severe Artifacts) & -67 & -86 & -72 & -225 \\
\midrule
\textbf{Final Curated Synthetic} & \textbf{36} & \textbf{21} & \textbf{35} & \textbf{92} \\
\bottomrule
\end{tabular}
\end{table}

\begin{table}[htbp]
\centering
\caption{Metadata Summary of the Final Evaluation Dataset. The statistics reflect the authentic captures and the inherent output configurations of the respective AI models used for the 92 curated fake videos.}
\label{tab:dataset_metadata}
\footnotesize
\begin{tabular}{lcccc}
\toprule
\textbf{Video Property} & \textbf{Authentic (Real)} & \textbf{Veo 3.1 Pro} & \textbf{Grok Imagine} & \textbf{LTX-2 Video} \\
\midrule
Video Count & 87 & 36 & 21 & 35 \\
Resolution & Mostly $1920\times1080$ & $1280\times720$ & $848\times480$ & $1920\times1080$ \\
Average FPS & 59.8 & 24.0 & 24.0 & 25.0 \\
Average Duration & 14.4 s & 8.0 s & 8.0 s & 10.3 s \\
\bottomrule
\end{tabular}
\end{table}

\subsection{Full Prompt Templates}
\label{appendix:prompt_templates}

This section provides the complete text of the prompt templates engineered for the adversarial video synthesis pipeline. For each of the three kinematic paradigms, we developed a prompt for the first-frame-conditioned Image-to-Video (I2V) generation, and a corresponding prompt for the pure Text-to-Video (T2V) generation. 

As discussed in Section~\ref{appendix:synthesis_stats}, the I2V prompts proved significantly more effective at preserving the required physical geometry of the grating structure across all tested models.

\subsubsection{Paradigm 1: Moving Camera, Static Subject}
This paradigm evaluates the generative model's ability to render natural optical flow and Moiré interference driven solely by camera translation, while maintaining the rigid structure of a stationary subject and grating. The corresponding prompts are detailed in \cref{fig:prompt1_i2v} and \cref{fig:prompt1_t2v}.

\vspace{1ex}
\noindent
\begin{minipage}{\linewidth}
\lstset{
  breaklines=true,
  basicstyle=\ttfamily\scriptsize,
  frame=single,
  columns=fullflexible
}
\begin{lstlisting}
Using the provided first frame, generate a photorealistic video of the same person clearly holding a rectangular moire board in their hands. The board is physically grasped by the person and remains in their hands for the entire video.

The moire board is a real, solid object with vertical black and white stripes printed on its surface. The stripes exist only on this physical board.

The person and the board remain completely stationary in the scene.
The board must stay fixed in the person's hands.
The board cannot move independently, cannot drift, cannot expand, cannot shrink, cannot rotate, and cannot deform.
Its size and position in the frame must remain consistent with the first frame.

Only the camera moves. The camera translates smoothly and strictly horizontally from left to right at constant speed.
No vertical motion. No forward or backward motion. No zoom. No tilt. No rotation.

The vertical stripes on the board must remain perfectly vertical.
Any visible moire interference must occur only on the surface of the board as a natural optical result of horizontal camera movement.
No stripes, patterns, or interference effects may appear anywhere outside the board.
Do NOT generate any new moving pattern in the background or across the room.
There must be no global animated overlay.

The background must remain stable and pattern-free.
Lighting, shadows, and perspective must stay physically realistic and consistent with the first frame.

Maintain sharp detail and strong temporal coherence across all frames.
Duration: 8 seconds. Frame rate: 24-30 fps. Style: natural documentary realism.
\end{lstlisting}
\captionof{figure}{I2V Prompt 1 (First-Frame Conditioned) used for Paradigm 1: Moving Camera, Static Subject.}
\label{fig:prompt1_i2v}
\end{minipage}

\vspace{2ex}
\noindent
\begin{minipage}{\linewidth}
\lstset{
  breaklines=true,
  basicstyle=\ttfamily\scriptsize,
  frame=single,
  columns=fullflexible
}
\begin{lstlisting}
Generate a photorealistic video from scratch of a person clearly holding a rectangular moire board in their hands. The board is physically grasped by the person and remains in their hands for the entire video.

The moire board is a real, solid object with vertical black and white stripes printed on its surface. The stripes exist only on this physical board.

The person and the board remain completely stationary in the scene.
The board must stay fixed in the person's hands.
The board cannot move independently, cannot drift, cannot expand, cannot shrink, cannot rotate, and cannot deform.
Its size and shape must remain constant throughout the entire video.

Only the camera moves. The camera translates smoothly and strictly horizontally from left to right at constant speed.
No vertical motion. No forward or backward motion. No zoom. No tilt. No rotation.

The vertical stripes on the board must remain perfectly vertical.
Any visible moire interference must occur only on the surface of the board as a natural optical result of horizontal camera movement.
No stripes, patterns, or interference effects may appear anywhere outside the board.
Do NOT generate any new moving pattern in the background or across the room.

The background must remain stable and pattern-free.
Lighting, shadows, and perspective must stay physically realistic.
Maintain sharp detail and strong temporal coherence across all frames.
Duration: 8 seconds. Frame rate: 24-30 fps. Style: natural documentary realism.
\end{lstlisting}
\captionof{figure}{T2V Prompt 1 (Text-Only) used for Paradigm 1: Moving Camera, Static Subject.}
\label{fig:prompt1_t2v}
\end{minipage}

\clearpage
\subsubsection{Paradigm 2: Static Camera, Moving Subject}
This condition challenges the model to synthesize smooth human motion while dynamically altering the Moiré effect based strictly on the subject's translation relative to a fixed viewpoint. The prompts are shown in \cref{fig:prompt2_i2v} and \cref{fig:prompt2_t2v}.

\vspace{1ex}
\noindent
\begin{minipage}{\linewidth}
\lstset{
  breaklines=true,
  basicstyle=\ttfamily\scriptsize,
  frame=single,
  columns=fullflexible
}
\begin{lstlisting}
Using the provided first frame, generate a photorealistic 8-second video of the same person clearly holding a rectangular moire board in their hands. The board is physically grasped by the person and remains in their hands for the entire video.

The moire board is a real, solid object with vertical black and white stripes printed on its surface. The stripes exist only on this physical board.

The camera remains completely stationary for the entire video.
Only the person moves.

The person moves smoothly and strictly horizontally (left or right) at a constant speed.
No vertical movement. No forward or backward movement.
No rotation of the body toward or away from the camera. No tilting of the board.

The board must stay fixed in the person's hands.
The board cannot move independently, cannot drift, cannot expand, cannot shrink, cannot rotate, and cannot deform.
Its size, orientation, and shape must remain consistent with the first frame.
The board must always face the camera directly and remain upright.

The vertical stripes on the board must remain perfectly vertical.
Any visible moire interference must occur only on the surface of the board as a natural optical result of horizontal movement.
No stripes, patterns, or interference effects may appear anywhere outside the board.
Do NOT generate any new moving pattern in the background.
No global animated overlay.

The background must remain stable and pattern-free.
Lighting, shadows, and perspective must stay physically realistic and consistent with the first frame.
Maintain sharp detail and strong temporal coherence across all frames.
Duration: 8 seconds. Frame rate: 24-30 fps. Style: natural documentary realism.
\end{lstlisting}
\captionof{figure}{I2V Prompt 2 (First-Frame Conditioned) used for Paradigm 2: Static Camera, Moving Subject.}
\label{fig:prompt2_i2v}
\end{minipage}

\vspace{2ex}
\noindent
\begin{minipage}{\linewidth}
\lstset{
  breaklines=true,
  basicstyle=\ttfamily\scriptsize,
  frame=single,
  columns=fullflexible
}
\begin{lstlisting}
Generate a photorealistic 8-second video of a person clearly holding a rectangular moire board in their hands. The board is physically grasped by the person and remains in their hands for the entire video.

The moire board is a real, solid object with vertical black and white stripes printed on its surface. The stripes exist only on this physical board.

The camera remains completely stationary for the entire video.
Only the person moves.

The person moves smoothly and strictly horizontally (left or right) at a constant speed.
No vertical movement. No forward or backward movement.
No rotation of the body toward or away from the camera. No tilting of the board.

The board must stay fixed in the person's hands.
The board cannot move independently, cannot drift, cannot expand, cannot shrink, cannot rotate, and cannot deform.
The board must always face the camera directly and remain upright.

The vertical stripes on the board must remain perfectly vertical.
Any visible moire interference must occur only on the surface of the board as a natural optical result of horizontal movement.
No stripes, patterns, or interference effects may appear anywhere outside the board.
Do NOT generate any new moving pattern in the background.
No global animated overlay.

The background must remain stable and pattern-free.
Lighting, shadows, and perspective must stay physically realistic.
Maintain sharp detail and strong temporal coherence across all frames.
Duration: 8 seconds. Frame rate: 24-30 fps. Style: natural documentary realism.
\end{lstlisting}
\captionof{figure}{T2V Prompt 2 (Text-Only) used for Paradigm 2: Static Camera, Moving Subject.}
\label{fig:prompt2_t2v}
\end{minipage}

\clearpage
\subsubsection{Paradigm 3: Simultaneous Movement}
This represents the most complex kinematic scenario, requiring the model to maintain the relative scale and perspective of the subject while simultaneously generating coherent background translation and optical interference. The prompts are provided in \cref{fig:prompt3_i2v} and \cref{fig:prompt3_t2v}.

\vspace{1ex}
\noindent
\begin{minipage}{\linewidth}
\lstset{
  breaklines=true,
  basicstyle=\ttfamily\scriptsize,
  frame=single,
  columns=fullflexible
}
\begin{lstlisting}
Using the provided first frame, generate a photorealistic 8-second video of the same person clearly holding a rectangular moire board in their hands. The board is physically grasped by the person and remains in their hands for the entire video.

The moire board is a real, solid object with vertical black and white stripes printed on its surface. The stripes exist only on this physical board.

Both the camera and the person move smoothly and strictly horizontally (left or right) at the same constant speed and in the same direction.

The relative position between the camera and the person remains constant.
There is no change in framing, scale, or perspective.
The person stays centered in the frame.

No vertical movement. No forward or backward movement. No zoom. No tilt. No rotation.

The board must stay fixed in the person's hands.
The board cannot move independently, cannot drift, cannot expand, cannot shrink, cannot rotate, and cannot deform.
Its size, orientation, and shape must remain identical to the first frame.
The board must always face the camera directly and remain upright.

The vertical stripes on the board must remain perfectly vertical.
Any visible moire interference must occur only on the surface of the board as a natural optical result of horizontal motion.
No stripes, patterns, or interference effects may appear anywhere outside the board.
Do NOT generate any new moving pattern in the background.
No global animated overlay.

The background may translate horizontally due to motion, but must remain physically realistic and stable.
Maintain sharp detail and strong temporal coherence across all frames.
Duration: 8 seconds. Frame rate: 24-30 fps. Style: natural documentary realism.
\end{lstlisting}
\captionof{figure}{I2V Prompt 3 (First-Frame Conditioned) used for Paradigm 3: Simultaneous Movement.}
\label{fig:prompt3_i2v}
\end{minipage}

\vspace{2ex}
\noindent
\begin{minipage}{\linewidth}
\lstset{
  breaklines=true,
  basicstyle=\ttfamily\scriptsize,
  frame=single,
  columns=fullflexible
}
\begin{lstlisting}
Generate a photorealistic 8-second video of a person clearly holding a rectangular moire board in their hands. The board is physically grasped by the person and remains in their hands for the entire video.

The moire board is a real, solid object with vertical black and white stripes printed on its surface. The stripes exist only on this physical board.

Both the camera and the person move smoothly and strictly horizontally (left or right) at the same constant speed and in the same direction.

The relative position between the camera and the person remains constant.
There is no change in framing, scale, or perspective.
The person stays centered in the frame.

No vertical movement. No forward or backward movement. No zoom. No tilt. No rotation.

The board must stay fixed in the person's hands.
The board cannot move independently, cannot drift, cannot expand, cannot shrink, cannot rotate, and cannot deform.
The board must always face the camera directly and remain upright.

The vertical stripes on the board must remain perfectly vertical.
Any visible moire interference must occur only on the surface of the board as a natural optical result of horizontal motion.
No stripes, patterns, or interference effects may appear anywhere outside the board.
Do NOT generate any new moving pattern in the background.
No global animated overlay.

The background may translate horizontally due to motion, but must remain physically realistic and stable.
Maintain sharp detail and strong temporal coherence across all frames.
Duration: 8 seconds. Frame rate: 24-30 fps. Style: natural documentary realism.
\end{lstlisting}
\captionof{figure}{T2V Prompt 3 (Text-Only) used for Paradigm 3: Simultaneous Movement.}
\label{fig:prompt3_t2v}
\end{minipage}
\section{Additional Experimental Results}
\label{appendix:results}

\subsection{Threshold-Based Real/Fake Classification}
\label{appendix:classification}

\vspace{1ex}
\noindent\textbf{Problem Setup and Notation.} 
Let each video sample be represented by a scalar score $s \in \mathbb{R}$, denoting the maximum correlation (\texttt{best\_correlation}) produced by our detector. We define the binary ground-truth labels as $y=1$ for authentic real videos (captured across all indoor and outdoor settings) and $y=0$ for AI-generated fake videos. Following our data curation protocol (detailed in \cref{appendix:attack}), generated samples that exhibited catastrophic physical failure (yielding exactly $s=0$) or invalid non-finite values (NaN/Inf) are excluded from this threshold analysis. These samples are trivially detectable by the naked eye and are therefore treated as invalid for continuous score-distribution analysis. 

Given a detection threshold $\tau$, the prediction rule is defined as:
\begin{equation}
\hat{y}(\tau) = 
\begin{cases} 
1, & \text{if } s \ge \tau \quad \text{(Predict Real)} \\
0, & \text{if } s < \tau \quad \text{(Predict Fake)}
\end{cases}
\end{equation}

\vspace{1ex}
\noindent\textbf{Evaluation Metrics.}
By sweeping the threshold $\tau$, we compute the standard components of the confusion matrix: True Positives (TP), False Negatives (FN), True Negatives (TN), and False Positives (FP). From these, we derive the True Positive Rate ($\text{TPR} = \frac{\text{TP}}{\text{TP} + \text{FN}}$) and True Negative Rate ($\text{TNR} = \frac{\text{TN}}{\text{TN} + \text{FP}}$).

We report both standard \textit{Accuracy} and \textit{Balanced Accuracy}. While standard Accuracy measures the overall proportion of correct predictions, it can be biased by class imbalance. Balanced Accuracy, defined as $\frac{1}{2}(\text{TPR}(\tau) + \text{TNR}(\tau))$, equally weights the positive (real) and negative (fake) recall, providing a more robust metric for imbalanced datasets.

\vspace{1ex}
\noindent\textbf{ROC Curve and Threshold Selection.}
The Receiver Operating Characteristic (ROC) curve is generated by plotting the TPR against the False Positive Rate ($\text{FPR} = 1 - \text{TNR}$) across all possible operational thresholds. The global separability of the two distributions, independent of any specific threshold, is summarized by the Area Under the Curve (AUC), approximated numerically via trapezoidal integration.

For practical deployment scenarios requiring a binary decision, we select the optimal threshold $\tau^*$ by maximizing the standard accuracy:
\begin{equation}
\tau^* = \arg\max_\tau \text{Accuracy}(\tau)
\end{equation}
If multiple thresholds yield identical accuracy, we break ties by selecting the one with the higher Balanced Accuracy.

\begin{figure}[t]
\centering
\begin{minipage}{0.48\linewidth}
  \centering
  \includegraphics[width=\linewidth]{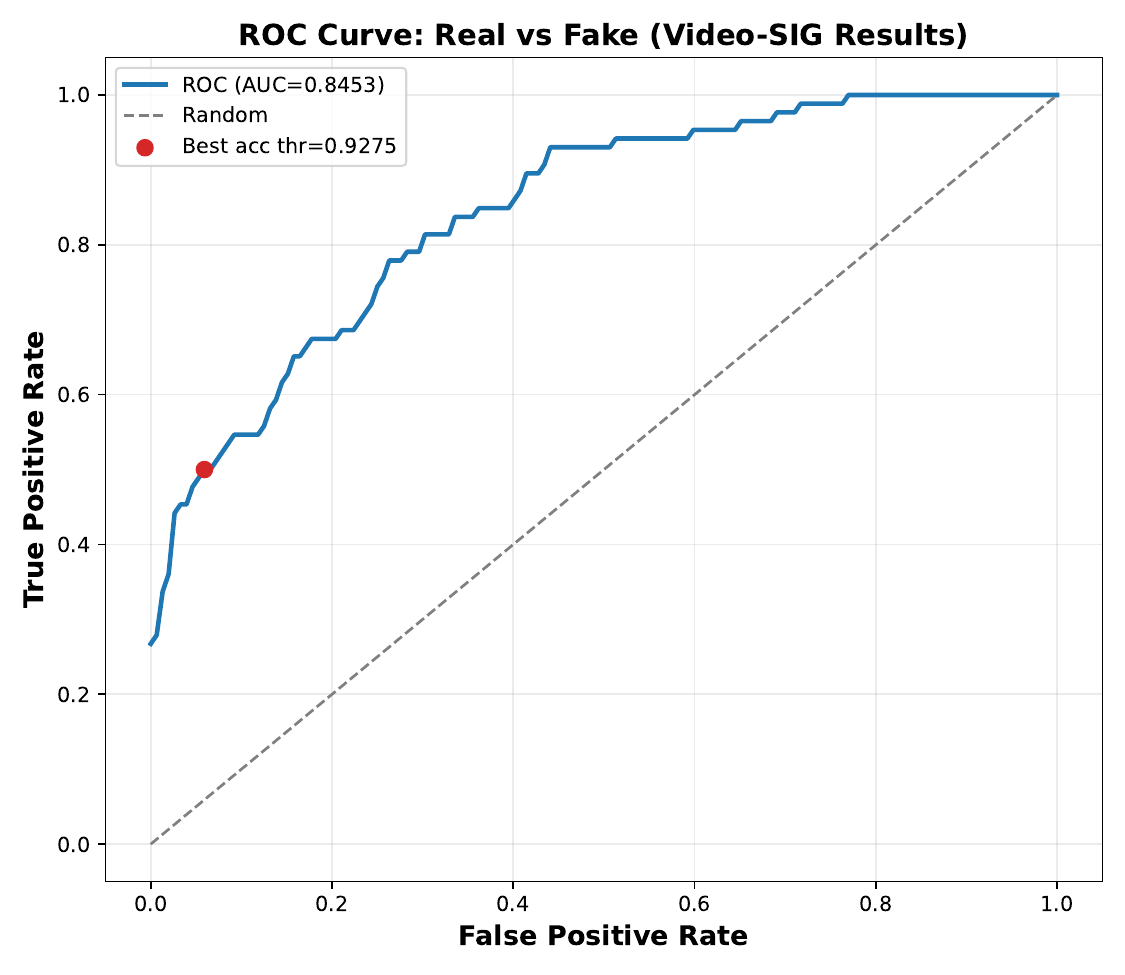}
  \caption{ROC Curve for real vs. fake video classification. Our method achieves an AUC of 0.8453, demonstrating strong global separability based on the Moiré correlation signature.}
  \label{fig:roc_curve}
\end{minipage}
\hfill
\begin{minipage}{0.48\linewidth}
  \centering
  \includegraphics[width=\linewidth]{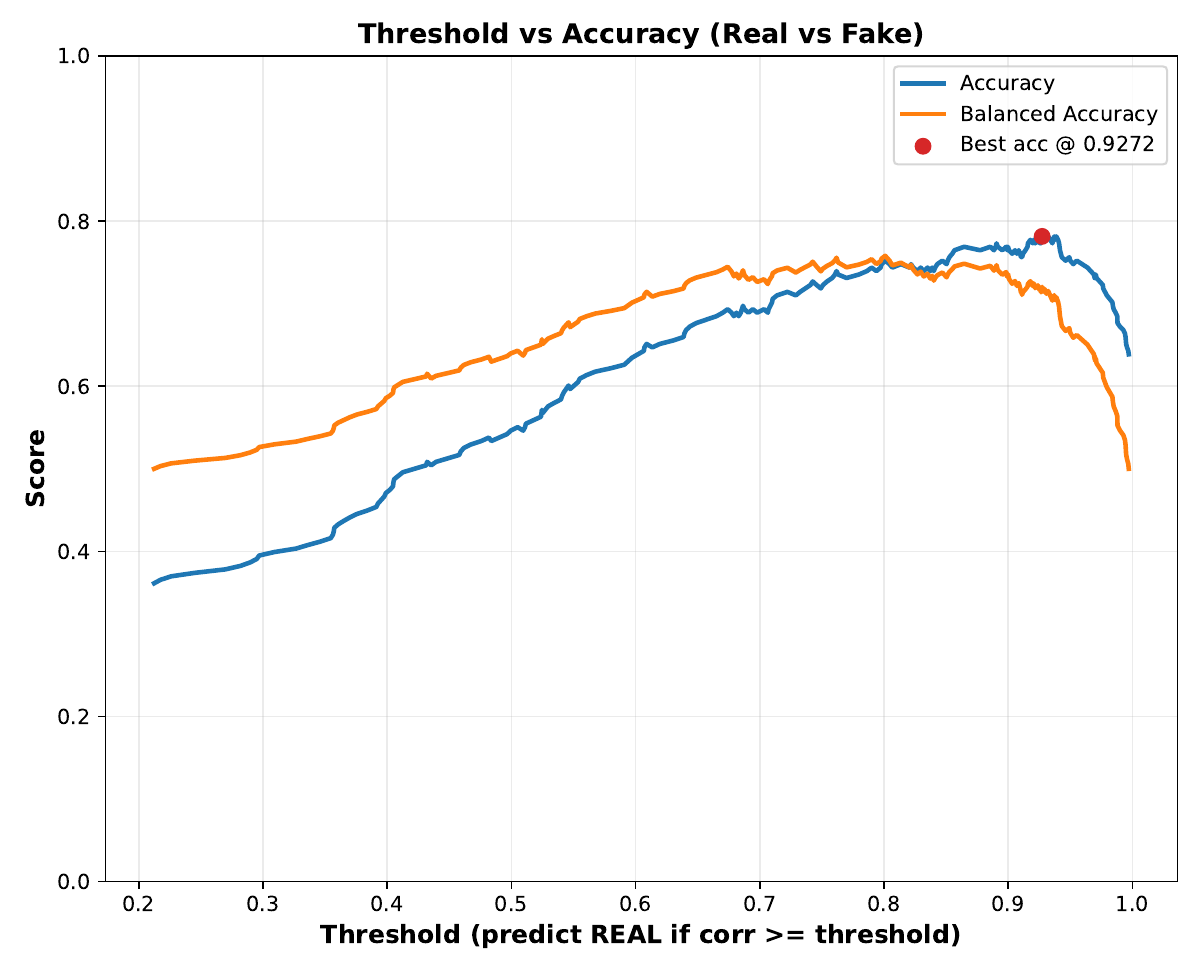}
  \caption{Impact of the decision threshold ($\tau$) on Accuracy and Balanced Accuracy. The optimal operational point is identified at $\tau^* \approx 0.927$.}
  \label{fig:threshold_acc}
\end{minipage}
\end{figure}

\vspace{1ex}
\noindent\textbf{Results and Analysis.}
\cref{fig:roc_curve} presents the ROC curve of our detection algorithm, achieving a strong AUC of $0.8453$. It is crucial to note that these metrics are computed \textit{exclusively} on the heavily curated subset of 92 high-quality adversarial videos (as detailed in \cref{appendix:synthesis_stats}), deliberately excluding 225 easily detectable generative failures. Consequently, this 0.8453 AUC reflects the algorithm's discriminative capability under a highly challenging, "hard-negative" evaluation setting, ensuring that the performance is not artificially inflated by trivially flawed generations.

Furthermore, as illustrated in \cref{fig:threshold_acc}, our threshold selection criterion identifies an optimal decision boundary at an exceptionally high correlation value of $\tau^* \approx 0.927$, where both standard and Balanced Accuracy reach their peak. The high value of this optimal threshold fundamentally underscores the strictness of our proposed Moiré motion invariant: authentic physical captures consistently maintain near-perfect phase-displacement correlations, whereas even the most visually deceptive AI generations fail to satisfy this rigid mathematical and optical constraint.

\subsection{Additional Robustness Tests on Pattern Layout and 3D Slant}
\label{appendix:geom_stress}

To further evaluate whether the proposed Moiré signature is tied to a single planar stripe layout, we conducted additional physics-based simulations under alternative pattern designs and more challenging board geometries. These tests complement the real-video and AI-generation experiments in the main paper by isolating geometric factors that are difficult to control precisely in physical capture.

\begin{figure}[t]
    \centering
    \includegraphics[width=\linewidth]{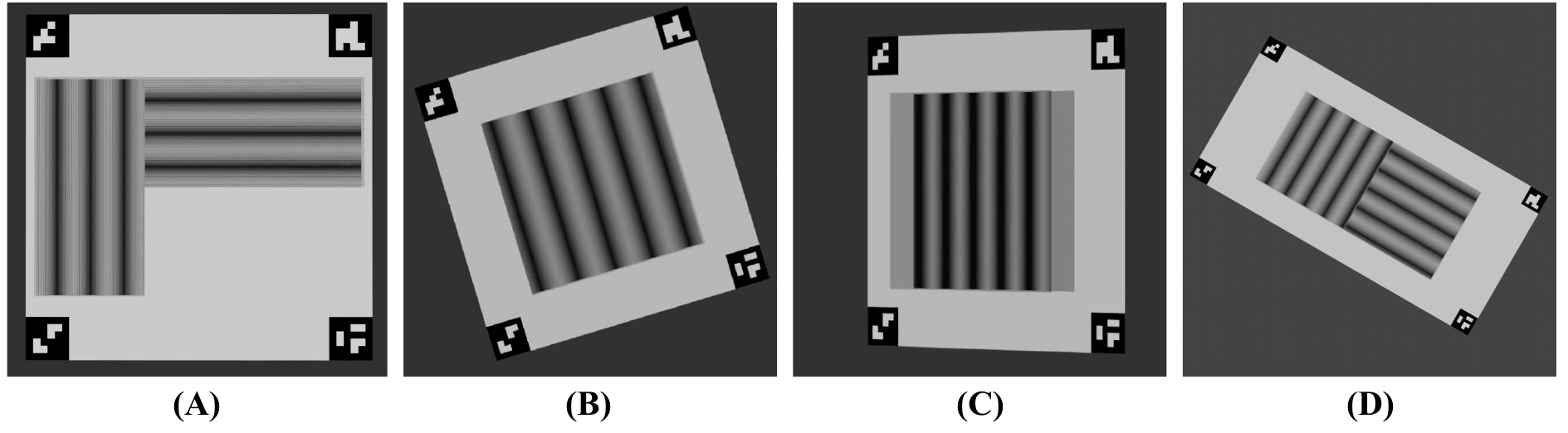}
    \caption{\textbf{Additional simulated robustness tests.}
    (A) Orthogonal multi-fringe regions for multi-directional sensitivity.
    (B) A rotated single-axis tag requiring rotation compensation.
    (C) A 45$^\circ$ viewing/motion stress test.
    (D) A dual-axis tag under compound board rotation.
    These simulations evaluate whether the Moiré motion signature remains measurable beyond the single fronto-parallel stripe configuration used in the main prototype.}
    \label{fig:geom_stress}
\end{figure}

\cref{fig:geom_stress} shows the four tested configurations. 
(A) uses two orthogonal Moiré fringe regions on the same board, allowing the verifier to extract independent phase-motion signals along two axes. 
(B) evaluates single-axis tags under in-plane board rotation, where rotation compensation is required before computing the correlation. 
(C) tests a 45$^\circ$ viewing/motion configuration, where the observed motion is not perfectly aligned with the fringe-sensitive axis. 
(D) combines two orthogonal fringe regions with compound board rotation, producing the most challenging setting among these simulations.

\paragraph{Metrics.}
For each simulated video, we extract the cumulative Moiré phase signal and the corresponding board-motion signal using the same correlation-based verifier as in the main paper. 
We report the following quantities in \cref{tab:geom_stress_results}:
\begin{itemize}
    \item \textbf{Region/axis}: the Moiré sub-region being evaluated and the motion axis to which that fringe orientation is primarily sensitive.
    \item \textbf{Rot. comp.}: whether rotation compensation is applied before measuring the motion signal.
    \item \textbf{Window mean $|\rho|$}: the mean absolute Pearson correlation over 30-frame sliding windows. This is our primary metric, since the authentication signal is evaluated over short temporal windows rather than individual frames.
    \item \textbf{Window std.}: the standard deviation of the window-level absolute correlations, measuring temporal stability.
    \item \textbf{Window range}: the minimum and maximum absolute window correlations across the video.
    \item \textbf{Global $|\rho|$}: the absolute Pearson correlation computed over the entire sequence. This is reported for completeness, but can be lower than the window-level score when the trajectory is non-monotonic or contains multiple motion phases.
    \item \textbf{Decision}: the detector output produced by the same correlation-based classification logic used in our pipeline.
\end{itemize}

\begin{table}[t]
\centering
\small
\setlength{\tabcolsep}{3.5pt}
\renewcommand{\arraystretch}{1.08}
\caption{\textbf{Additional robustness results under alternative pattern layouts and 3D slant.}
We report the absolute Pearson correlation between Moiré phase displacement and board motion. 
The primary metric is the mean absolute 30-frame window correlation. 
For multi-axis tags, each fringe region is evaluated separately.}
\label{tab:geom_stress_results}
\resizebox{\linewidth}{!}{
\begin{tabular}{cllccccc}
\toprule
\textbf{Panel} & \textbf{Setting} & \textbf{Region / Axis} & \textbf{Rot. comp.} & \textbf{Frames} & \textbf{Window mean $|\rho|$} & \textbf{Window range} & \textbf{Global $|\rho|$} \\
\midrule
A & Orthogonal multi-fringe tag & Vertical / $x$ & No & 180 & 0.951 & [0.722, 0.998] & 0.807 \\
A & Orthogonal multi-fringe tag & Horizontal / $y$ & No & 180 & 0.854 & [0.561, 0.999] & 0.325 \\
\midrule
B & Rotated single-axis tag, trajectory 1 & $x$ & Yes & 120 & 0.758 & [0.033, 0.990] & 0.141 \\
B & Rotated single-axis tag, trajectory 2 & $x$ & Yes & 120 & 0.809 & [0.445, 0.987] & 0.024 \\
\midrule
C & 45$^\circ$ viewing/motion stress test & $x$ & Yes & 120 & 0.720 & [0.139, 0.997] & 0.447 \\
\midrule
D & Dual-axis tag with compound rotation & Vertical / $x$ & Yes & 240 & 0.667 & [0.045, 0.998] & 0.713 \\
D & Dual-axis tag with compound rotation & Horizontal / $y$ & Yes & 240 & 0.570 & [0.004, 0.988] & 0.709 \\
\bottomrule
\end{tabular}}
\end{table}

\paragraph{Analysis.}
The results show that the Moiré motion signature remains measurable across different stripe orientations, board layouts, and slanted viewing conditions. 
The orthogonal multi-fringe configuration achieves strong window-level correlations on both axes, supporting the feasibility of multi-directional patterns for practical deployments where the user or camera motion may not align with a single fringe-sensitive direction. 
The rotated single-axis and 45$^\circ$ stress tests further indicate that moderate in-plane rotation and oblique motion weaken but do not eliminate the phase-motion coupling when rotation compensation is applied.

The hardest case is the dual-axis tag under compound rotation. 
As expected, this setting introduces stronger axis cross-talk, perspective distortion, and local contrast variation, leading to lower and less stable window-level correlations. 
Nevertheless, both axes still provide measurable correlation signals, and the combined output remains classified as likely real by the detector. 
These results suggest that 3D slant primarily reduces signal-to-noise ratio rather than invalidating the underlying physical coupling. 
In practical systems, such degradation can be mitigated by multi-axis fusion, window-level reliability filtering, and rejection of frames where the rectified fringe contrast is insufficient for stable phase estimation.
\section{Discussion and Future Work}
\label{appendix:discussion}

While our proof-of-concept validates the Moiré motion invariant and demonstrates its effectiveness against state-of-the-art video generative models, there remain important limitations and avenues for future research before deployment at scale. We discuss these aspects and outline potential system-level solutions.

\vspace{1ex}
\noindent\textbf{Toward Real-World Deployment and Multi-directional Patterns.}
Currently, our empirical evaluation demonstrates strong discriminative performance in controlled laboratory and standard indoor/outdoor settings. However, large-scale, "in-the-wild" deployment will require robustness against extreme environmental factors, such as severe motion blur, low-light noise, and aggressive social media compression (e.g., H.264/H.265 transcoding on platforms like WhatsApp or X). 
The additional simulations in \cref{appendix:geom_stress} suggest that \textit{multi-directional Moiré patterns} are a promising direction for strengthening the signature against such degradation. A one-dimensional grating produces fringes sensitive to camera motion along a single axis. Incorporating multiple grating orientations (e.g., orthogonal or hexagonal grids) within a single assembly extends sensitivity to arbitrary motion trajectories. This not only ensures a strong correlation signal regardless of how the user moves the camera but also exponentially increases the dimensionality of the authentication signature. A generative model would need to simultaneously reproduce the correct phase-displacement coupling along every grating axis, compounding the difficulty for an attacker.


\vspace{1ex}
\noindent\textbf{Advanced Generative Attacks and Fundamental Limits.}
Our threat model assumes that current generative architectures (Diffusion and Transformers) learn statistical mappings rather than performing explicit physics-based simulations of Moiré optics. While we utilized meticulously optimized prompts to generate the strongest possible adversarial videos (as detailed in \cref{appendix:attack}), a highly sophisticated attacker might attempt to bypass the system by rendering the Moiré board using a traditional 3D graphics engine (e.g., Unreal Engine with full ray-tracing) and compositing it into an AI-generated scene. While theoretically possible, achieving photorealistic coherence between a ray-traced object and a generative background—including lighting, shadows, and perspective matching—requires immense manual effort, thereby neutralizing the primary threat of "effortless, large-scale AI generation." 
Finally, our method strictly requires relative motion between the camera and the grating assembly to produce a measurable phase-displacement signal. Authentication is fundamentally impossible if both the camera and the subject remain entirely static throughout the recording. Future software interfaces must require the user to perform a slight scanning motion to trigger the verification protocol.

\end{document}